\documentclass[11pt]{article}

\usepackage[preprint]{acl}

\usepackage{times}
\usepackage{latexsym}
\usepackage{fontawesome5}
\usepackage{tabularx}
\usepackage{svg}

\usepackage[T1]{fontenc}

\usepackage[utf8]{inputenc}

\usepackage{microtype}

\usepackage{inconsolata}

\usepackage{graphicx}

\usepackage{amsmath}
\usepackage{booktabs}
\usepackage[skins]{tcolorbox}
\usepackage{multicol}
\usepackage{tcolorbox}
\tcbuselibrary{breakable}
\usepackage{cuted}
\usepackage{dblfloatfix}
\usepackage{amssymb}
\usepackage{multirow}

\usepackage{pifont}
\newcommand{\cmark}{\ding{51}}   
\newcommand{\xmark}{\ding{55}}   
\newcommand{\pmark}{$\circ$}     

\title{AffectWorld: Benchmarking Emotion-Driven Future World Prediction}

\author{
  Changhao Song\textsuperscript{1},
  Yacong Chen\textsuperscript{2},
  Hui Gao\textsuperscript{1},
  Yazhou Zhang\textsuperscript{1}\thanks{Corresponding author.},
  Peng Zhang\textsuperscript{1}\thanks{Corresponding author.}
  \\\\
  \textsuperscript{1}School of Computer Science and Technology, Tianjin University, Tianjin, China \\
  \textsuperscript{2}International Joint Institute of Tianjin University, Fuzhou, China \\
  \small{\texttt{\{songchanghao, yc\_chen, hui\_gao, yzhou\_zhang, pzhang\}@tju.edu.cn}}
}

\begin{document}
\maketitle
\begin{abstract}
The world encompasses both physical environments and human actors. Predicting its evolution requires understanding both physical laws and how emotions shape human actions. However, existing world model benchmarks primarily assess physical plausibility and perceptual quality, leaving emotion-driven world evolution largely untested.
We introduce AffectWorld, a benchmark that evaluates the chain Emotion → Action → World State through three tasks: World-State Consistency, Emotion–Action Consistency, and Causal Consistency. It comprises 1,498 everyday video samples annotated with plausible emotions, action tendencies, and atomic world-state constraints. 
A counterfactual subset of 223 samples tests whether changing an agent's emotion under fixed scene and event conditions yields corresponding changes in actions and consequences. 
We define seven metrics across the three tasks and an additional counterfactual metric, combining automatic visual assessment with human-calibrated VLM/LLM judgments. Evaluations of eight SoTA world models (e.g., Cosmos 3) required world-state change scores of 0.064–0.240 and counterfactual emotion sensitivity scores of 0.027–0.070 on a [0, 1] scale. These results reveal substantial gaps in generating the consequences of emotion-driven actions and adapting future predictions to emotional interventions. AffectWorld provides a testbed for advancing world models that capture how emotions shape future physical and social states\footnote{https://anonymous.4open.science/r/AffectWorld-Benchmark-370C/}.
\end{abstract}


\begin{figure*}
    \centering
    \includegraphics[width=1.0\textwidth]{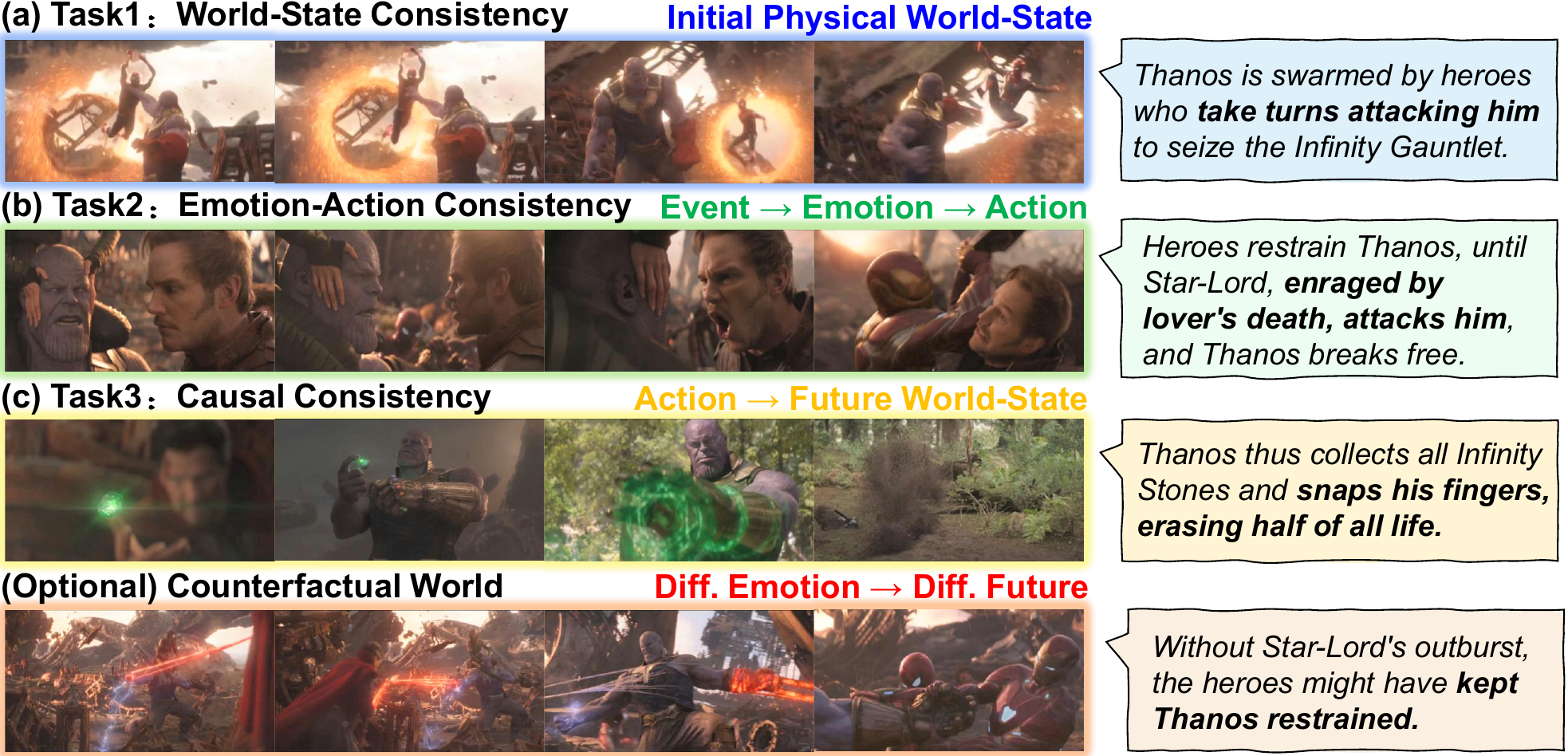}
\caption{Illustration of the AffectWorld task design, using scenes from \textit{Avengers: Infinity War} as an example.
The causal chain is initial world state $\rightarrow$ event $\rightarrow$ emotion $\rightarrow$ action $\rightarrow$ future world state.
And the counterfactual world represents a different emotion leading to a different future.}
\label{fig:bench_pipeline}
\end{figure*}

\section{Introduction}
Imagine a person dining who receives distressing news and accidentally drops the cup in their hand: the cup shatters on the floor, fragments scatter, and a nearby diner steps back in alarm. This sequence follows physical laws while also being shaped by human emotion. From its Old English etymological meaning, ``the age of man''~\citep{zhang2026toward}, to modern dictionary definitions encompassing the Earth, its inhabitants and everything upon it~\citep{etymonlineWorld,cambridgeWorld}, the concept of the ``world'' has a distinct human dimension. We understand the world as an interconnected whole comprising the natural environment and human society, whose evolution is constrained by physical laws and human emotional dynamics. World models (WMs) predict future states conditioned on observations and actions, providing a foundation for agent planning and decision-making~\cite{ding2025understanding}. Hence, evaluating world models requires addressing a question: \textit{can these models predict how emotions influence human actions and how those actions subsequently change the physical and social environment?}

Recently, WMs evaluation has expanded beyond visual quality and temporal continuity to encompass physical plausibility and embodied capabilities. For physical commonsense, VideoPhy~\cite{bansal2025videophy} examines whether object interactions in generated videos obey physical laws. WorldBench~\cite{upadhyay2026worldbench} provides concept-specific diagnostics and evaluates measurable parameters such as gravity and friction. For embodied applications, WorldSimBench~\cite{qin2024worldsimbench} evaluates video generation models as world simulators, EWMBench~\cite{yue2025ewmbench} assesses scene consistency, motion correctness and semantic alignment, WorldEval~\cite{li2025worldeval} examines world models as robot policy evaluators, and World-in-World~\cite{zhang2026world} measures their utility through performance in closed-loop tasks. 

However, these benchmarks lack evaluations of how emotions drive future changes in the world. Generating scenes that obey physical laws does not necessarily demonstrate an ability to predict why a person takes a particular action or how that action changes the surrounding environment. World model benchmarks must evaluate the connections between emotion, action and world-state change: whether WMs can generate contextually plausible emotions and behaviours together with their physical and social consequences, and whether predicted actions and consequences change appropriately when the emotion changes.

We introduce AffectWorld, a benchmark for emotion-driven future world prediction covering solo, dyadic and group scenarios. AffectWorld comprises 1,498 daily video samples, takes historical video observations as input, and provides fine-grained annotations of event context, plausible emotions, action tendencies and world-state constraints. As illustrated in Figure~\ref{fig:bench_pipeline}, the benchmark follows a causal chain from the initial world state through event, emotion and action to the future world state. It organizes evaluation into three progressive tasks: \textbf{(1) World-State Consistency}, which assesses whether the social world state evolves plausibly and coherently, \textbf{(2) Emotion–Action Consistency}, which examines whether emotions are plausible given the event context and whether behaviours are compatible with those emotions, and \textbf{(3) Causal Consistency}, which tests whether actions produce plausible changes in world state.

AffectWorld accommodates multiple plausible futures: the same world can follow different trajectories depending on how people feel and act~\citep{merton1968selffulfilling,rosenthal1968pygmalionurban}. Building on this view, different emotional states can favor different actions and lead to different physical and social outcomes.
The benchmark also includes a counterfactual subset of 223 samples. By intervening only on the target individual's emotion while holding the physical scene and event context fixed, this subset tests whether different emotions produce different but equally plausible futures.

We define eight core metrics: \textit{identity preservation}, \textit{spatial structure stability} and \textit{temporal coherence} for Task 1, \textit{event–emotion plausibility} and \textit{emotion–action consistency} for Task 2, and \textit{must, must-not, directional} world-state changes for Task 3. An additional counterfactual emotion sensitivity metric assesses whether actions and consequences respond appropriately to changes in emotion. Evaluation combines automatic perceptual metrics with human-calibrated VLM/LLM judgments.

We evaluate eight SoTA models: Cosmos 3, Wan2.1-VACE, LTX-Video, HunyuanVideo-1.5, CogVideoX, Stable Video Diffusion, DynamiCrafter and EasyAnimate. The results reveal substantial capability gaps: required world-state change scores reach only 0.064-0.240, counterfactual emotion sensitivity scores range from 0.027 to 0.070, and emotion-action consistency under counterfactual emotions drops by 48\%.. Although the models can maintain physical consistency, they struggle to infer plausible emotions, translate those emotions into appropriate actions, and generate the resulting consequences. Performance declines from dyadic to group settings for 7/8 models on Task~1 and 8/8 on Task~2. LTX-Video's scores drop by 17.9\% and 9.5\%, highlighting the challenge of maintaining world-state consistency and emotion-aligned behavior in group scenes. AffectWorld provides a unified benchmark for evaluating and improving world models' ability to predict emotion-driven human behaviour and its consequences.

Our contributions are: 
\begin{itemize}
    \item We introduce AffectWorld to benchmark emotion-driven future world prediction.
    \item We provide 1,498 annotated videos, including 223 counterfactual samples.
    \item We develop nine evaluation metrics and identify capability gaps across eight models.
\end{itemize}

\begin{table*}[h]
\centering
\setlength{\tabcolsep}{4pt}
\renewcommand{\arraystretch}{0.86}
\setlength{\aboverulesep}{1pt}
\setlength{\belowrulesep}{1pt}
\caption{Comparison of representative world model benchmarks. \cmark, \pmark, and \xmark denote full, partial, and no coverage.}
\label{tab:benchmark_comparison}
\small
\begin{tabular*}{\textwidth}{@{\extracolsep{\fill}}lccccc@{}}
\toprule
\textbf{Benchmark}
& \textbf{Scene Consistency}
& \textbf{World Dynamics}
& \textbf{instruction}
& \textbf{Embodied}
& \textbf{Affect Causality} \\
\midrule


VBench~\cite{huang2024vbench}
& \cmark & \xmark & \xmark & \xmark & \xmark \\

EvalCrafter~\cite{liu2024evalcrafter}
& \cmark & \xmark & \xmark & \xmark & \xmark \\

EWMBench~\cite{yue2025ewmbench}
& \cmark & \pmark & \xmark & \xmark & \xmark \\

\midrule


VideoPhy~\cite{bansal2025videophy}
& \pmark & \cmark & \xmark & \xmark & \xmark \\

WorldBench~\cite{upadhyay2026worldbench}
& \xmark & \cmark & \xmark & \xmark & \xmark \\

WorldScore~\cite{duan2025worldscore}
& \cmark & \cmark & \cmark & \xmark & \xmark \\

\midrule


WorldSimBench~\cite{qin2024worldsimbench}
& \pmark & \pmark & \cmark & \cmark & \xmark \\

WorldEval~\cite{li2025worldeval}
& \xmark & \xmark & \xmark & \cmark & \xmark \\

World-in-World~\cite{zhang2026world}
& \xmark & \pmark & \cmark & \cmark & \xmark \\

WorldArena~\cite{shang2026worldarena}
& \cmark & \pmark & \cmark & \cmark & \xmark \\

\midrule

\textbf{AffectWorld (Ours)}
& \textbf{\cmark}
& \textbf{\cmark}
& \textbf{\cmark}
& \textbf{\pmark}
& \textbf{\cmark} \\

\bottomrule
\end{tabular*}
\end{table*}

\section{Related Work}

\subsection{World Models}
World model research has expanded from learning predictive representations to simulating physical environments and, more recently, modeling emotional and social dynamics.

\textbf{Prediction-centred world models.}
Early world models learned compact representations of environmental dynamics to predict future states and support decision-making~\cite{ha2018world}. Latent dynamics models, exemplified by Dreamer, enabled agents to learn policies through imagined trajectories~\cite{hafner2023mastering}. This line established state-transition prediction as a foundation for planning and policy learning.

\textbf{Physical world models.}
Video-based approaches extended prediction to observable scenes, generating future frames and action-conditioned trajectories. Sora explored large-scale video generation as a route towards world simulation~\cite{brooks2024video}, Genie introduced interactive environments conditioned on actions~\cite{bruce2024genie}, and Cosmos developed world foundation models for physical AI~\cite{agarwal2025cosmos}. These capabilities support data synthesis~\cite{jang2025dreamgen}, policy evaluation~\cite{li2025worldeval} and action planning~\cite{hu2024video}, with an emphasis on physical dynamics and task execution.

\textbf{Emotional world models.}
Emotional world modeling extends the predictive state to include the internal factors that shape human behaviour. LEWM~\cite{song2025large} pioneered emotion-driven multimodal world modeling by explicitly representing emotional states and predicting future world states conditioned on their emotional evolution.
Related approaches address social and mental dynamics: Social World Models represent implicit intentions, beliefs and evolving social states~\cite{zhou2025social}, while Mental World Modeling jointly simulates physical and mental state transitions~\cite{fei2026mental}. These developments motivate evaluating whether emotional states translate into observable actions and subsequent changes.

Although emotional world models have emerged, evaluation has not kept pace with their modeling objectives. Representing or predicting emotions alone does not establish whether emotions influence subsequent behaviour. Dedicated benchmarks are therefore needed to test whether different emotional states lead to corresponding changes in actions and their physical and social consequences.

\subsection{World Model Benchmarks}
As world models have progressed toward predictive and interactive simulation, their evaluation has expanded across three dimensions: perceptual and semantic consistency, physical plausibility and dynamics, and embodied utility and interaction.

\textbf{Perceptual and Semantic Benchmarks.}
VBench evaluates video generation across visual quality, temporal consistency, motion quality, and semantic alignment~\cite{huang2024vbench}. EvalCrafter provides a comprehensive evaluation of large video generation models, covering visual quality, temporal coherence, motion, and semantic consistency~\cite{liu2024evalcrafter}. EWMBench extends evaluation to embodied world models by measuring scene consistency, motion correctness, and semantic quality~\cite{yue2025ewmbench}. Together, these benchmarks assess the visual coherence and semantic consistency of generated scenes.

\textbf{Physical Dynamics Benchmarks.}
VideoPhy evaluates physical commonsense through interactions involving solids and fluids~\cite{bansal2025videophy}, while WorldBench isolates individual physical concepts such as gravity, friction, and viscosity~\cite{upadhyay2026worldbench}. WorldScore further evaluates controllability, generation quality, and dynamics across different world-generation settings~\cite{duan2025worldscore}. These benchmarks focus on physical constraints and the controllable evolution of generated worlds.

\textbf{Embodied Utility Benchmarks.}
WorldSimBench combines perceptual assessment with video-to-action evaluation in embodied environments~\cite{qin2024worldsimbench}. WorldEval examines world models as evaluators of real-world robot policies~\cite{li2025worldeval}, while World-in-World evaluates their utility in closed-loop tasks~\cite{zhang2026world}. WorldArena further integrates perceptual evaluation with synthetic data generation, policy evaluation, and action planning~\cite{shang2026worldarena}. These works evaluate the use of world models for action, planning, policy evaluation, and interaction.

Despite this progress, existing benchmarks provide limited evaluation of emotion-driven world evolution. AffectWorld combines world-state consistency with explicit evaluation of emotion-action compatibility and action consequences. It uses state constraints to accommodate multiple plausible futures and paired emotional conditions to test whether changing emotion produces corresponding changes in the predicted future. Table~\ref{tab:benchmark_comparison} compares AffectWorld with existing benchmarks.

\section{AffectWorld Benchmark}
\label{sec:benchmark}

\begin{figure*}[h]
    \centering
    \includegraphics[width=1.0\textwidth]{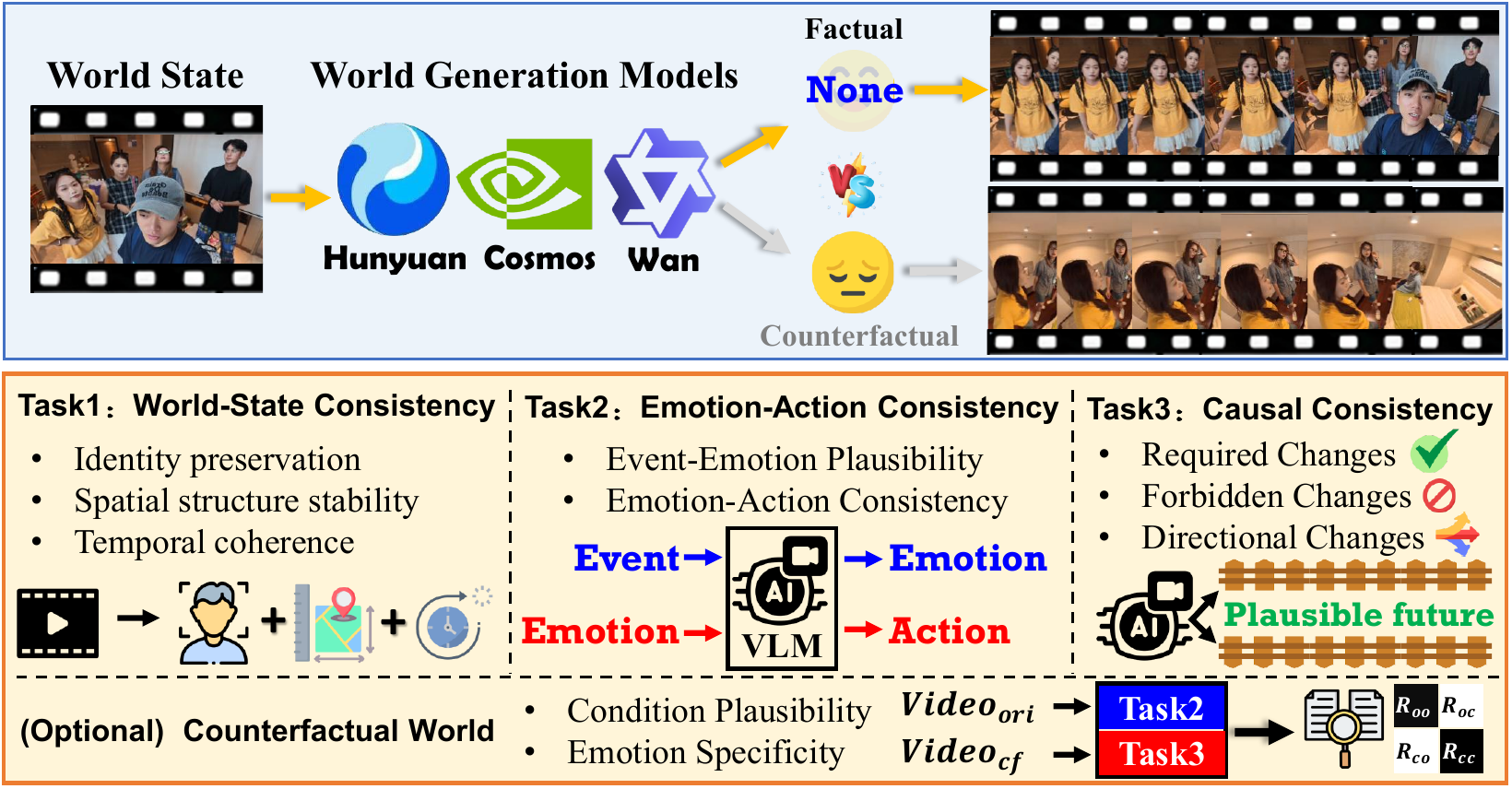}
    \caption{Overview of the AffectWorld generation and evaluation process.}
    \label{fig:ewh_bench_pipeline}
\end{figure*}

We introduce AffectWorld, a unified benchmark designed to evaluate whether a model can treat emotion as a latent causal variable and predict the coupled physical-social future, as shown in Fig.~\ref{fig:ewh_bench_pipeline}.

\subsection{Data Construction}
\label{subsec:dataset}

\subsubsection{Data Acquisition}
\label{subsubsec:data_collection}

We collected videos of daily scenes from Douyin and TikTok published during the preceding three to four years. We obtained 7,136 candidate clips from 675 source videos, covering individual activities, interpersonal interactions, and group settings.

We retained videos that met three criteria: a identifiable target person whose identity remained consistent, observable emotional cues and an identifiable event context, and subsequent emotion-related behaviour or world-state changes. We excluded videos with identity discontinuities, cuts that disrupted the event-response sequence, insufficient evidence of emotion or event context, or no observable subsequent behaviour or state change. Specifically, source videos were segmented into 5–15-second clips using shot-boundary detection and adaptive windows. Using the temporal annotations described in Section~\ref{subsubsec:data_annotation}, we extracted $state_0$ as the historical observation and retained $state_1$ as the response reference. Observation clips were encoded as H.264 MP4, preserving source frame rates and aspect ratios with a height cap of 1,280 pixels. Available audio was retained as AAC, separate subtitle and data streams were removed.

\begin{figure}[h]
    \centering
    \includegraphics[width=1.0\linewidth]{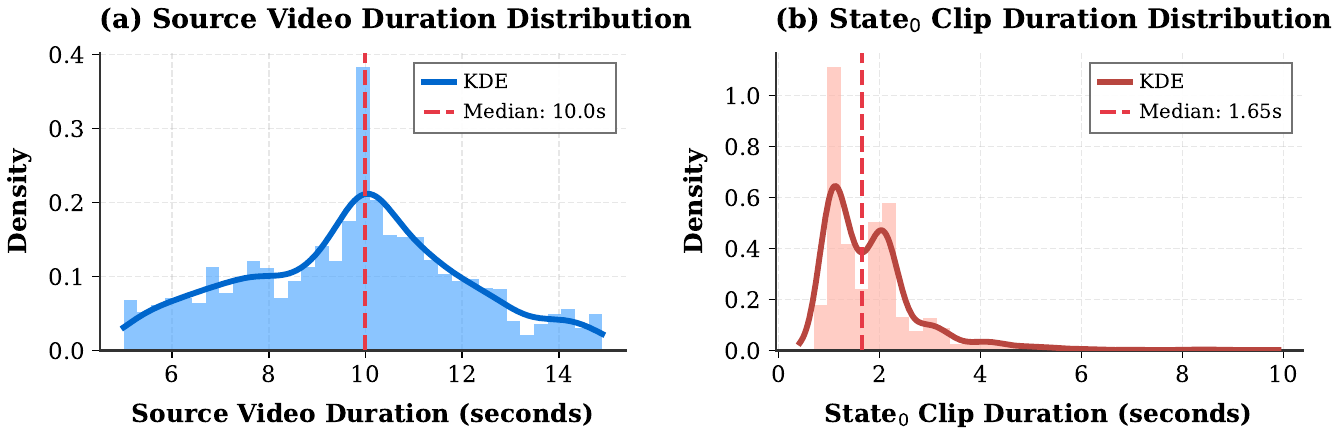}
    \caption{Duration distributions of (a) source videos and (b) \texttt{state\_0} clips in the benchmark dataset.}
    \label{fig:duration_dist}
\end{figure}

The final dataset contains 1,498 source clips with a mean duration of $9.8 \pm 2.3$ seconds (mean $\pm$ standard deviation) and a median of $10.0$ seconds. Historical observations $state_0$ have a mean duration of $1.86 \pm 1.05$ seconds and a median of $1.65$ seconds. Figure~\ref{fig:duration_dist} shows both duration distributions.

\subsubsection{Data Annotation and Quality Control}
\label{subsubsec:data_annotation}

We construct the annotations through a two-stage pipeline. A VLM first extracts temporally anchored visual evidence, and a LLM converts these descriptions into structured records containing scene, emotion, action, and world-state information. Annotation prompts and are provided in Appendix~\ref{app:annotation}. We then construct the counterfactual subset and apply automatic and human quality-control procedures.

\begin{table*}[h]
\centering
\small
\setlength{\tabcolsep}{4pt}
\renewcommand{\arraystretch}{1.08}
\caption{Annotation taxonomies for scenes, emotions, and action
tendencies.}
\label{tab:taxonomies}

\begin{tabularx}{\textwidth}{@{}l l X@{}}
\toprule
\textbf{Taxonomy} & \textbf{Category / Attribute} & \textbf{Labels} \\
\midrule

\multirow{4}{*}{\textbf{Scene}}
  & Scenario category
  & Daily Life, Personal Hobby, Romantic, Education, Travel,
    Workplace, Healthcare, Personal Service, Consultation,
    Family Life, Friend Social, Sports, Leisure, Public Transit,
    Community, Shopping \\
  & Interaction configuration
  & Solo (1 participant), Dyadic (2 participants),
    Group ($\geq 3$ participants) \\

\midrule

\multirow{6}{*}{\textbf{Emotion}}
  & Sadness
  & Guilty, Despair, Lonely, Bored, Disappointed, Awful \\
  & Fear
  & Scared, Anxious, Insecure, Rejected \\
  & Anger
  & Mad, Hurt, Threatened, Distant \\
  & Surprise
  & Confused, Startled, Amazed, Excited \\
  & Happiness
  & Joyful, Proud, Optimistic, Peaceful, Loving, Hopeful \\
  & Disgust
  & Avoidance, Hesitant, Aversion, Judgmental, Revolted,
    Disapproval \\

\midrule

\multirow{7}{*}{\textbf{Action}}
  & Approach-Avoidance
  & Approach, Avoidance, Freeze \\
  & Stress Response
  & Escape, Appeasement \\
  & Emotion Regulation
  & Emotional Expression, Emotional Suppression, Self-Calming \\
  & Social Interaction
  & Social Engagement, Withdrawal, Support Seeking,
    Support Providing \\
  & Conflict Dynamics
  & Confrontation, Escalation, De-escalation \\
  & Attention \& Cognition
  & Attention Shift, Information Seeking \\
  & Task \& Environment
  & Object Interaction, Task Persistence, Disengagement \\

\bottomrule
\end{tabularx}
\end{table*}

\textbf{VLM-based visual evidence extraction.}
We use Qwen-3.8-Max to generate chronological descriptions of the source videos. The model tracks the same target person and records the initial scene, events, immediate responses, actions, and others' reactions. Temporal markers identify key stages, typically yielding 4-10 markers per video. Descriptions report observable evidence, such as looking down, or stopping a conversation, without assigning emotions, intentions or causal explanations.

\textbf{LLM-based structured annotation.}
Then we use GPT-5.6-sol to convert the visual descriptions into JSON records. Each record contains the scene category, historical interval $state_0$, initial emotion, event context, target emotion $e_{target}$, alternative plausible emotions $plausible\_emotions$, plausible action tendencies $plausible\_actions$ and world-state constraints. 

\begin{figure*}[h]
    \centering
    \includegraphics[width=1.0\textwidth]{category_wheels.pdf}
    \caption{Taxonomy and distribution of (a) scene, (b) emotions, and (c) action tendencies.}
    \label{fig:category}
\end{figure*}

\textbf{Annotation taxonomy.}
Specifically, the scene, emotion and action fields are annotated using predefined taxonomies. 
We annotate scenes along two separate axes: \textit{scenario category}, which describes the activity or context using 16 labels, and \textit{interaction configuration}, which distinguishes Solo, Dyadic, and Group settings by the number of participants involved in the focal event (one, two, and three or more, respectively), excluding incidental bystanders. Scenario categories are not restricted to a fixed interaction configuration. 
The emotion taxonomy comprises 30 subcategories under six basic categories, following basic-emotion and hierarchical emotion taxonomies~\cite{ekman1992argument,shaver1987emotion}. 
The action taxonomy comprises 20 tendencies across seven categories, grounded in emotional
action readiness~\cite{frijda1986emotions,frijda1989relations}.
Table~\ref{tab:taxonomies} lists the complete taxonomies.

\textbf{World-state constraints.}
To accommodate multiple plausible futures, we characterize world-state changes along four dimensions: (1) physical object states (position, attributes, and existence), (2) spatial relations (distance, orientation, and contact), (3) agent states (expression, posture, speech, and attention), and (4) social atmosphere (conversation rhythm and interaction patterns).
For each sample, we annotate only constraints strongly causally related to its event and emotion. Each constraint item is associated with one state dimension and belongs to one of three types: required changes ($must$), which every plausible future should contain; forbidden changes ($must\_not$), which no plausible future should contain; or directional
changes ($directional$), which specify a trend without requiring an exact target state. A sample may contain one or more constraint types and cover one or more state dimensions. These constraints characterize shared properties of plausible futures without imposing a single
reference trajectory.

\textbf{Counterfactual subset construction.}
For samples admitting a plausible alternative emotion, we select a counterfactual emotion $e_{c}$ from the emotion taxonomy, where $e_{c} \neq e_{o}$ and $e_{o}$ is the annotated target emotion. The characters, physical scene, and event context are held fixed. Annotators then derive the plausible actions and world-state constraints associated with $e_{c}$, producing original and counterfactual branches that share $state_0$. The alternative emotion must fit the existing context, and its associated actions and effects must be plausible under that condition. The resulting subset contains 223 samples, each providing paired emotion conditions and evaluation constraints.

\textbf{Quality control.}
We select 300 original and 60 counterfactual cases through stratified random sampling across scene modes and emotion categories for independent review by two annotators. Reviewers check person identity and temporal correspondence, visual evidence for emotional interpretations, action plausibility, and agreement between world-state constraints and observation evidence. For counterfactual cases, they check that the fixed conditions remain unchanged and that the alternative emotion, actions and consequences are compatible. Agreement on the single-label target emotion is measured using Cohen's $\kappa$, yielding $\kappa = 0.85$. 

\subsubsection{Dataset Analysis}
\label{subsubsec:data_analysis}

\textbf{Scene and event composition.}
AffectWorld contains 1,498 video samples covering 16 everyday scenarios (Figure~\ref{fig:category}a). Solo, Dyadic and Group settings account for 44.0\%, 40.9\% and 15.2\% of the samples, respectively. The three most frequent scenarios are Daily Life, Education and Leisure, together accounting for 47.8\% of the dataset. The least frequent scenario is Healthcare, representing 0.9\%. A qualitative visualization of the 16 scenario categories is provided in Appendix~\ref{app:analysis}. The main event themes include other agent's action, object event, environmental event, and task outcome, with other agent's action being the most frequent, accounting for 37.5\% of the samples.

\textbf{Major categories and long-tailed distributions.}
Each annotated branch may carry several plausible emotions and action tendencies. Based on plausible emotion labels, the three most frequent emotions are joyful, anxious and excited, which together appear in 66.9\% of the annotated branches (Figure~\ref{fig:category}b). The least frequent emotion that appears is distant, with only two branches. Based on the number of branches containing each action tendency, the three most frequent actions are emotional expression, object interaction and social engagement (Figure~\ref{fig:category}c), appearing in 50.1\%, 39.2\% and 37.5\% of the branches, respectively. There are 15 emotion categories and 5 action categories represented in fewer than 30 branches, indicating long-tailed category distributions.

\begin{figure}[!htbp]
    \centering
    \includegraphics[width=1.0\linewidth]{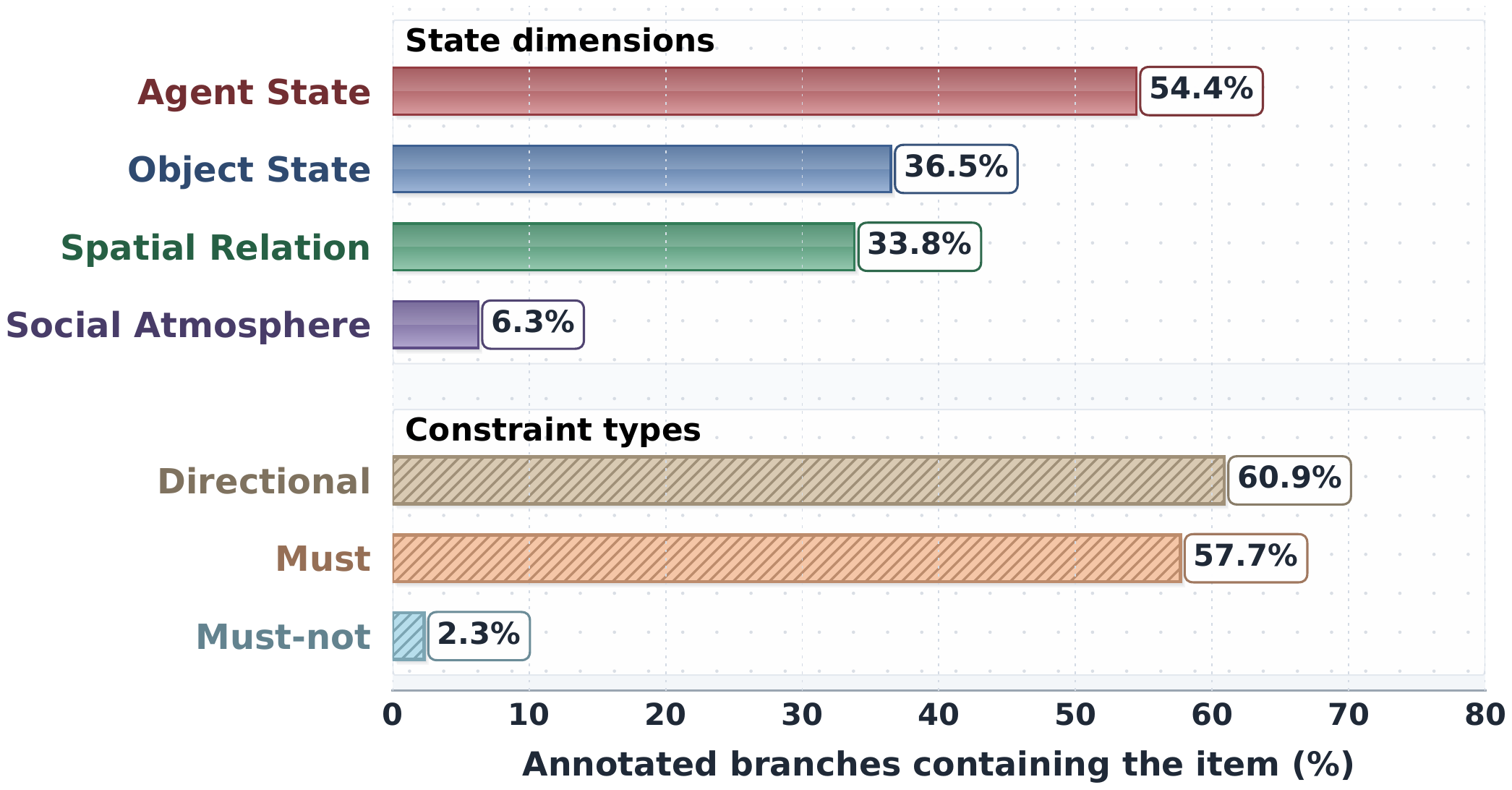}
    \caption{Proportion of samples containing each state dimension (top) and constraint type (bottom).}
    \label{fig:constraints}
\end{figure}

\textbf{Candidate emotions, actions and world-state constraints.}
Each sample contains an average of 1.32 plausible emotions, 2.25 plausible actions and 1.48 world-state constraints. Samples with multiple plausible emotions and those with multiple plausible actions account for 27.3\% and 83.3\% of the dataset, respectively. As shown in Figure~\ref{fig:constraints}, constraints on physical object states, spatial relations, agent states and social atmosphere appear in 36.5\%, 33.8\%, 54.4\% and 6.3\% of the 1,721 annotated branches, comprising 1,498 original and 223 counterfactual branches. Among these branches, 28.2\% involve at least two state dimensions. Branches containing $must$, $must\_not$ and $directional$ constraints account for 57.7\%, 2.3\% and 60.9\%.

\subsection{Task 1: World-State Consistency}
\label{subsec:task1}

Task 1 assesses whether a continuation maintains consistent entities, scene geometry, and motion over time. These properties provide the visual foundation for evaluating emotion-driven actions. We measure three aspects: \textbf{identity preservation}, \textbf{spatial structure stability}, and \textbf{temporal coherence}. Details are provided in Appendix~\ref{app:task1_formulation}.

\textbf{Identity preservation.}
We use SAM2~\cite{ravi2025sam} to track salient agents and objects across frames, and measure the cosine similarity of their DINOv2~\cite{oquab2023dinov2} features between consecutive frames. The similarity is weighted by optical-flow magnitude to emphasize regions with meaningful motion.

\textbf{Spatial structure stability.}
We extract depth maps from each frame using Depth-Anything V2~\cite{yang2024depth} and quantify the temporal variance. Lower depth variation corresponds to a higher spatial stability score.

\textbf{Temporal coherence.}
We use VFI-Mamba~\cite{zhang2407vfimamba} to interpolate intermediate frames and compare them with the corresponding ground-truth frames using SSIM~\cite{wang2004image}. We then aggregate these similarities using optical-flow-based weights, assigning greater importance to midpoints with larger local motion.

The final Task 1 score is the geometric mean of the three component scores:
\begin{equation}
\mathrm{Score}_{T1}
=
\sqrt[3]{S_{id} \cdot S_{depth} \cdot S_{temp}}.
\label{eq:scoret1}
\end{equation}

\subsection{Task 2: Emotion-Action Consistency}
\label{subsec:task2}

Task 2 evaluates whether a model has learned the transition from event and emotion to observable behavior. We decompose it into two successive dimensions: \textbf{Event-Emotion Plausibility}, which tests whether the generated future exhibits a contextually plausible emotion, and \textbf{Emotion-Action Consistency}, which tests whether the inferred emotion drives a compatible observable behavior.

Both dimensions are scored with a two-stage judge that decouples perception from judgment. A VLM first converts the generated future video $V_{pred}$ into a time-anchored description $D = \mathrm{VLM}(V_{pred})$ of observable cues. An LLM judge $J$ then scores each candidate against $D$ on a graded rubric mapped to $[0,1]$. The event context $\mathcal{C}$ is used to interpret the reported cues.



For Event-Emotion plausibility, the judge matches the emotional tone described in $D$ against the candidate emotion set $\mathcal{E} = \{e_{target}\} \cup plausible\_emotions$:
\begin{equation}
M_{emo} = \max_{e \in \mathcal{E}} J_{emo}(D, e, \mathcal{C})
\label{eq:memo}
\end{equation}
where $e_{target}$ is the target emotion, and \textit{plausible\_emotions} is the set of plausible alternatives. To accommodate diverse emotional responses, we take the maximum score over the candidate set rather than enforcing a single target. We define the corresponding inferred emotion as
\begin{equation}
e^* = \arg\max_{e \in \mathcal{E}} J_{emo}(D, e, \mathcal{C}).
\label{eq:estar}
\end{equation}
Thus, $M_{emo}$ measures the best emotional match, while $e^*$ is used for the subsequent action-consistency evaluation. This dimension measures whether the generated future expresses a cognitively plausible emotional response to the event.

For Emotion-Action consistency, the judge evaluates whether the observable behavior in $D$ is compatible with the inferred emotion $e^*$. We allow multiple plausible action tendencies and take the maximum score over the candidate action set $\mathcal{A} = plausible\_actions$:
\begin{equation}
S_{action} = \max_{a \in \mathcal{A}} J_{real}(D, a, \mathcal{C}) \cdot w\big(J_{coh}(D, a, e^*)\big)
\label{eq:saction}
\end{equation}
where $J_{real}$ measures whether the observable behavior realizes the core function of the candidate action tendency $a$, and $J_{coh}$ measures whether the manner of that behavior, such as its speed, force, and bodily tone, is consistent with $e^*$. The function $w(\cdot)$ is a monotone gate that assigns zero to behaviors whose manner conflicts with $e^*$. 

The Task 2 score is the mean of two dimensions:
\begin{equation}
\mathrm{Score}_{T2} = \frac{1}{2}\left(M_{emo} + S_{action}\right).
\label{eq:scoret2}
\end{equation}

\subsection{Task 3: Causal Consistency}
\label{subsec:task3}

Task 3 evaluates whether a model captures the causal evolution
\textit{Emotion} $\rightarrow$ \textit{Action} $\rightarrow$ \textit{World State}. The evaluation focuses on the social and physical consequences of behavior, such as others' reactions and changes in interpersonal relationships.

Since an event admits many plausible futures, we annotate world-state constraints derived from the event, emotion, and action tendencies. These constraints characterize what all plausible futures share (required changes, \textit{must}), what none of them contain (forbidden changes, \textit{must-not}), and the directions in which the world should evolve (directional changes, \textit{directional}). An LLM judge $J$ checks these constraints against the semantic description $D$ extracted by a VLM from the generated video $V_{pred}$ and outputs graded scores in $[0,1]$.

The causal consistency of $V_{pred}$ comprises four components. $S_{must}$ is computed as the product of the scores for all required changes, whereas $S_{not}$ is the product of the non-violation scores for all forbidden changes. $S_{dir}$ averages the scores of the directional changes, and $S_{exists}$ is a binary indicator that verifies whether the generated future contains at least one world-state change. Detailed formulations of these components are provided in Appendix~\ref{app:task3_formulation}. We combine them as:
\begin{equation}
S_{causal}
=
S_{not}
\cdot
\frac{
S_{exists} + K_{must} S_{must} + K_{dir} S_{dir}
}{
1 + K_{must} + K_{dir}
}.
\label{eq:scausal}
\end{equation}
Here, $K_{must}$ and $K_{dir}$ denote the numbers of required and directional changes, respectively. Task 3 score is the average score over all $N$ samples:
\begin{equation}
\mathrm{Score}_{T3}
=
\bar{S}_{causal}
=
\frac{1}{N}
\sum_{i=1}^{N}
S_{causal}^{i}.
\label{eq:scoret3}
\end{equation}

\subsection{Optional Subtask: Counterfactual World}
\label{subsec:counterfactual}

To test whether a model treats emotion as a causal driver, we construct a curated counterfactual subset in which the scene, characters, and event context are held fixed while only the target agent's emotion is changed. Each sample defines two branches sharing $state_0$: $(state_0, e_{o} \rightarrow state_1^{o})$ and $(state_0, e_{c} \rightarrow state_1^{c})$, where $e_{o}$ and $e_{c}$ denote the original and counterfactual emotions.

Given $state_0$, the model generates $V_{o}$ under $e_{o}$ and $V_{c}$ under $e_{c}$. We reuse the action-consistency score from Task 2 and the causal-consistency score from Task 3 to compute condition plausibility as $R(V,e) = S_{action}(V,e) \cdot S_{causal}(V,e)$. Both scores use the specified emotion $e$, together with its annotated action tendencies and world-state constraints, in particular, $S_{action}$ uses $e$ rather than the inferred emotion $e^*$.

We evaluate both videos under both emotion conditions and denote the resulting scores by $R_{xy} = R(V_x,e_y)$, where $x,y \in \{o,c\}$ index the original and counterfactual conditions. For each emotion $e_y$, specificity $\mathrm{Spe}_y$ measures the relative drop in plausibility when its annotations are applied to the other video $V_{\bar{y}}$ rather than its own video $V_y$. This drop is normalized by $R_{yy}$ and clipped to $[0,1]$, with specificity set to zero when $R_{yy}=0$. Detailed formulations are provided in Appendix~\ref{app:counterfactual_formulation}.

The counterfactual sensitivity score combines the geometric mean of the two own-condition plausibility scores with their average specificity:
\begin{equation}
S_{counter}
=
\sqrt{R_{oo}R_{cc}}
\cdot
\frac{\mathrm{Spe}_{o}+\mathrm{Spe}_{c}}{2}.
\label{eq:scounter}
\end{equation}
This score rewards plausible futures under both emotion conditions and their specificity to the respective emotions. The global score averages $S_{counter}$ over the $N_{c}$ samples:
\begin{equation}
\bar{S}_{counter}
=
\frac{1}{N_{c}}
\sum_{i=1}^{N_{c}} S_{counter}^{i}.
\label{eq:barscounter}
\end{equation}

\newcolumntype{N}{>{\centering\arraybackslash}p{1.2cm}}
\newcolumntype{W}{>{\centering\arraybackslash}p{1.6cm}}
\newlength{\tasktwoheadwidth}
\newlength{\tasktwocolwidth}
\newcolumntype{T}{>{\centering\arraybackslash}p{\tasktwocolwidth}}

\begin{table*}[t]
    \centering
    \setlength{\tabcolsep}{5pt}
    \renewcommand{\arraystretch}{1.12}

    \settowidth{\tasktwoheadwidth}
        {\textbf{Task 2: Emotion-Action Consistency}}
    \setlength{\tasktwocolwidth}
        {\dimexpr(\tasktwoheadwidth-4\tabcolsep)/3+1pt\relax}

    \caption{Main results on AffectWorld under video-to-video (V2V)
    and image-to-video (I2V) input settings.}
    \label{tab:main_results}

    \resizebox{\textwidth}{!}{%
    \begin{tabular}{ll cccc TTT ccc}
        \toprule
        \multirow{2}{*}{\textbf{Input}}
          & \multirow{2}{*}{\textbf{Model}}
          & \multicolumn{4}{c}{\textbf{Task 1: World-State Consistency}}
          & \multicolumn{3}{c}{\textbf{Task 2: Emotion-Action Consistency}}
          & \multicolumn{3}{c}{\textbf{Task 3: Causal Consistency}} \\
        \cmidrule(lr){3-6}
        \cmidrule(lr){7-9}
        \cmidrule(lr){10-12}
        & & $S_{id}$ & $S_{depth}$ & $S_{temp}$
          & $\mathbf{Score}_{T1}$
          & $M_{emo}$ & $S_{action}$ & $\mathbf{Score}_{T2}$
          & $S_{must}$ & $S_{dir}$ & $\mathbf{Score}_{T3}$ \\
        \midrule

        \multirow{3}{*}{\textbf{V2V}}
        & Cosmos3
          & 0.8261 & 0.3227 & 0.7165 & 0.5089
          & 0.4131 & 0.5157 & 0.4644
          & 0.0981 & 0.2097 & 0.5414 \\
        & Wan2.1-VACE
          & 0.6799 & 0.2977 & 0.7324 & 0.4307
          & 0.4541 & 0.6107 & 0.5324
          & 0.1384 & 0.3286 & 0.5911 \\
        & LTX-Video
          & 0.9245 & 0.4009 & 0.8481 & 0.6275
          & 0.4547 & 0.5613 & 0.5080
          & 0.1620 & 0.2965 & 0.5270 \\

        \midrule

        \multirow{5}{*}{\textbf{I2V}}
        & HunyuanVideo-1.5
          & 0.7787 & 0.1167 & 0.6476 & 0.3069
          & 0.6488 & 0.7296 & \textbf{0.6892}
          & 0.2397 & 0.3591 & \textbf{0.6199} \\
        & CogVideoX
          & 0.9525 & 0.5452 & 0.8984 & \textbf{0.7380}
          & 0.5068 & 0.5873 & 0.5471
          & 0.1519 & 0.2565 & 0.5533 \\
        & SVD
          & 0.8267 & 0.2982 & 0.6573 & 0.5058
          & 0.4603 & 0.5379 & 0.4991
          & 0.0816 & 0.1970 & 0.5217 \\
        & DynamiCrafter
          & 0.8265 & 0.2223 & 0.6762 & 0.4362
          & 0.3773 & 0.4885 & 0.4329
          & 0.0643 & 0.1921 & 0.5570 \\
        & EasyAnimate
          & 0.8446 & 0.4363 & 0.7605 & 0.6113
          & 0.4582 & 0.5636 & 0.5109
          & 0.0958 & 0.2216 & 0.5348 \\

        \bottomrule
    \end{tabular}%
    }
\end{table*}

\section{Experiments}
\label{sec:experiments}

\subsection{Experimental Setup}
\label{sec:experimental_setup}

\textbf{Baselines.}
We evaluate eight SoTA WMs on all 1,498 AffectWorld samples. Cosmos3~\cite{agarwal2026cosmos}, Wan2.1-VACE~\cite{wan2025wan} and LTX-Video~\cite{HaCohen2024LTXVideo} receive the history video $state_0$. HunyuanVideo-1.5~\cite{kong2024hunyuanvideo}, Stable Video Diffusion~\cite{blattmann2023stable}, DynamiCrafter~\cite{xing2024dynamicrafter}, EasyAnimate~\cite{xu2024easyanimate}  and CogVideoX~\cite{yang2024cogvideox} receive its first frame together with a textual description of the scene and event context. 

\textbf{Generation settings.}
For each sample, we generate one continuation of approximately 5–6 seconds at model-specific resolutions (480p–720p) and frame rates (8, 16, or 24 fps), using deterministic per-sample seeds. On the 223 counterfactual samples, we generate paired continuations under the original and alternative emotion conditions, keeping the input and event context fixed.

\textbf{Evaluation settings.}
Task 1 uses the automatic perceptual metrics defined in Section~\ref{subsec:task1}. For semantic evaluation, Qwen-3.8-Max extracts observable cues from generated videos, and GPT-5.6-sol scores these descriptions against the annotations. Each judgment is repeated three times and aggregated using the median. Model parameters and human calibration are detailed in Appendix~\ref{app:evaluation_parameters} and Appendix~\ref{app:human_calibration}, respectively.

\subsection{Main Results}
\label{sec:main_result}

As shown in Table~\ref{tab:main_results}, Task~1 reveals a consistent imbalance among its three components. Spatial structure stability $S_{depth}$ is the lowest-scoring component
for all models. CogVideoX achieves identity preservation $S_{id}=0.9525$ and temporal coherence $S_{temp}=0.8984$, but its $S_{depth}$ reaches only 0.5452. SVD exhibits a larger gap between $S_{id}=0.8267$ and $S_{depth}=0.2982$. The repeated shortfall suggests that \textbf{current models preserve recognizable entities more reliably than the spatial structure surrounding them, leaving scene geometry a common weakness.}

Task~2 shows that models struggle more with generating plausible emotional responses than emotion-compatible actions. Event-Emotion plausibility $M_{emo}$ is lower than
Emotion-Action consistency $S_{action}$ for all models. Wan2.1-VACE scores 0.4541 on $M_{emo}$ vs. 0.6107 on $S_{action}$, even HunyuanVideo-1.5, which leads both dimensions, scores 0.6488 and 0.7296, respectively. CogVideoX's strength in Task~1 also does not translate into a lead on Task~2, where it scores 0.5471 compared with HunyuanVideo-1.5's 0.6892. These results suggest that \textbf{models can generate emotion-compatible behavior
without achieving equally strong contextual plausibility.}

\begin{table*}[h]
    \small
    \centering
    \setlength{\tabcolsep}{4pt}
    \renewcommand{\arraystretch}{1.12}
    \caption{Counterfactual emotion sensitivity on AffectWorld.}
    \label{tab:counterfactual_results}
    \resizebox{\textwidth}{!}{%
    \begin{tabular}{ll cccccccccc}
        \toprule
        \textbf{Input} & \textbf{Model}
          & $S_{action}^{o}$ & $S_{action}^{c}$
          & $S_{causal}^{o}$ & $S_{causal}^{c}$
          & $R_{oo}$ $\uparrow$ & $R_{oc}$ $\downarrow$
          & $R_{co}$ $\downarrow$ & $R_{cc}$ $\uparrow$
          & $\mathrm{Spe}_{avg}$ $\uparrow$
          & $\boldsymbol{\bar{S}_{counter}}$ \\
        \midrule

        \multirow{3}{*}{\textbf{V2V}}
        & Cosmos3
          & 0.4664 & 0.3246 & 0.4992 & 0.5193
          & 0.2471 & 0.1263 & 0.2997 & 0.1840
          & 0.2442 & 0.0529 \\
        & Wan2.1-VACE
          & 0.5852 & 0.3077 & 0.5251 & 0.5392
          & 0.3233 & 0.1272 & 0.3311 & 0.1726
          & 0.2630 & 0.0544 \\
        & LTX-Video
          & 0.4678 & 0.2791 & 0.4324 & 0.4512
          & 0.2580 & 0.1166 & 0.2848 & 0.1534
          & 0.2946 & 0.0467 \\

        \midrule

        \multirow{5}{*}{\textbf{I2V}}
        & HunyuanVideo-1.5
          & 0.6219 & 0.3255 & 0.5703 & 0.5053
          & 0.3738 & 0.1394 & 0.3203 & 0.1955
          & 0.2694 & \textbf{0.0704} \\
        & CogVideoX
          & 0.5084 & 0.1903 & 0.5044 & 0.3327
          & 0.2855 & 0.1295 & 0.1621 & 0.0786
          & 0.2428 & 0.0265 \\
        & SVD
          & 0.4795 & 0.2452 & 0.4755 & 0.4502
          & 0.2383 & 0.1374 & 0.2152 & 0.1214
          & 0.1935 & 0.0294 \\
        & DynamiCrafter
          & 0.4924 & 0.2486 & 0.4762 & 0.4973
          & 0.2509 & 0.1097 & 0.2403 & 0.1316
          & 0.2410 & 0.0413 \\
        & EasyAnimate
          & 0.5572 & 0.2413 & 0.4736 & 0.4841
          & 0.2779 & 0.1183 & 0.2668 & 0.1295
          & 0.2671 & 0.0387 \\

        \bottomrule
    \end{tabular}%
    }
\end{table*}

Task~3 extends this weakness to the consequences of behavior. HunyuanVideo-1.5 achieves Emotion-Action consistency $S_{action}=0.7296$, but its required changes $S_{must}$ reaches only 0.2397. DynamiCrafter similarly scores 0.4885 on $S_{action}$ but only 0.0643 on $S_{must}$. Directional consistency $S_{dir}$ is comparatively stronger, averaging 0.2576 across models and reaching 0.3591 for HunyuanVideo-1.5, although these values remain low. The scores suggest that \textbf{current models capture how the world might change only partially and struggle to generate the specific outcomes that must follow from emotion-driven actions.}
\textit{We omit $S_{not}$ from the main table because only 3 original samples contain must-not constraints, and $S_{not}=1$ for all valid evaluations, leaving the aggregate scores unchanged.}

Taken together, weaknesses in spatial stability, emotional plausibility, and consequence generation reveal that \textbf{current world models struggle to maintain the full chain from an event to an appropriate emotional response, a compatible action, and its effects on the world.}

\subsection{Counterfactual emotion sensitivity}

Table~\ref{tab:counterfactual_results} shows consistently low counterfactual emotion sensitivity, with even HunyuanVideo-1.5 reaching only $\bar{S}_{counter}=0.0704$. The component scores reveal two sources of this weakness. Across models, Emotion-Action consistency $S_{action}$ drops by 48\% under counterfactual emotions, while mean
causal consistency $S_{causal}$ changes from 0.495 to 0.472. The sharper behavioral decline is accompanied by a drop in mean own-condition plausibility from $R_{oo}=0.282$ to $R_{cc}=0.146$. Emotion specificity is also limited, averaging $\mathrm{Spe}_{avg}=0.252$. These results suggest that \textbf{models struggle both to produce appropriate behavior under the alternative emotion and to make the resulting future specific to that intervention.}

V2V models show more consistent performance across all baselines: all three outperform 4/5 I2V models, with mean counterfactual scores of 0.0513 vs. 0.0413. One possible explanation is that video histories provide motion and interaction cues that help constrain plausible continuations, whereas a single image leaves these dynamics less specified. HunyuanVideo-1.5 achieves the highest score despite using I2V inputs, but its $S_{action}$ still falls from 0.6219 under the original emotion to 0.3255 under the alternative emotion. Thus, \textbf{richer context may support plausible continuation, but adapting behavior to a changed emotional condition remains a challenge across both input settings.}

Counterfactual prediction is a core capability of world models. However, our results show that this capability remains weak for emotion-driven behavior. \textbf{There is therefore substantial room for future in predicting how emotional interventions change actions and their consequences.}

\subsection{Performance Across Scene Complexity}
\label{sec:scene_analysis}
\begin{figure}[h]
    \centering
    \includegraphics[width=\linewidth]{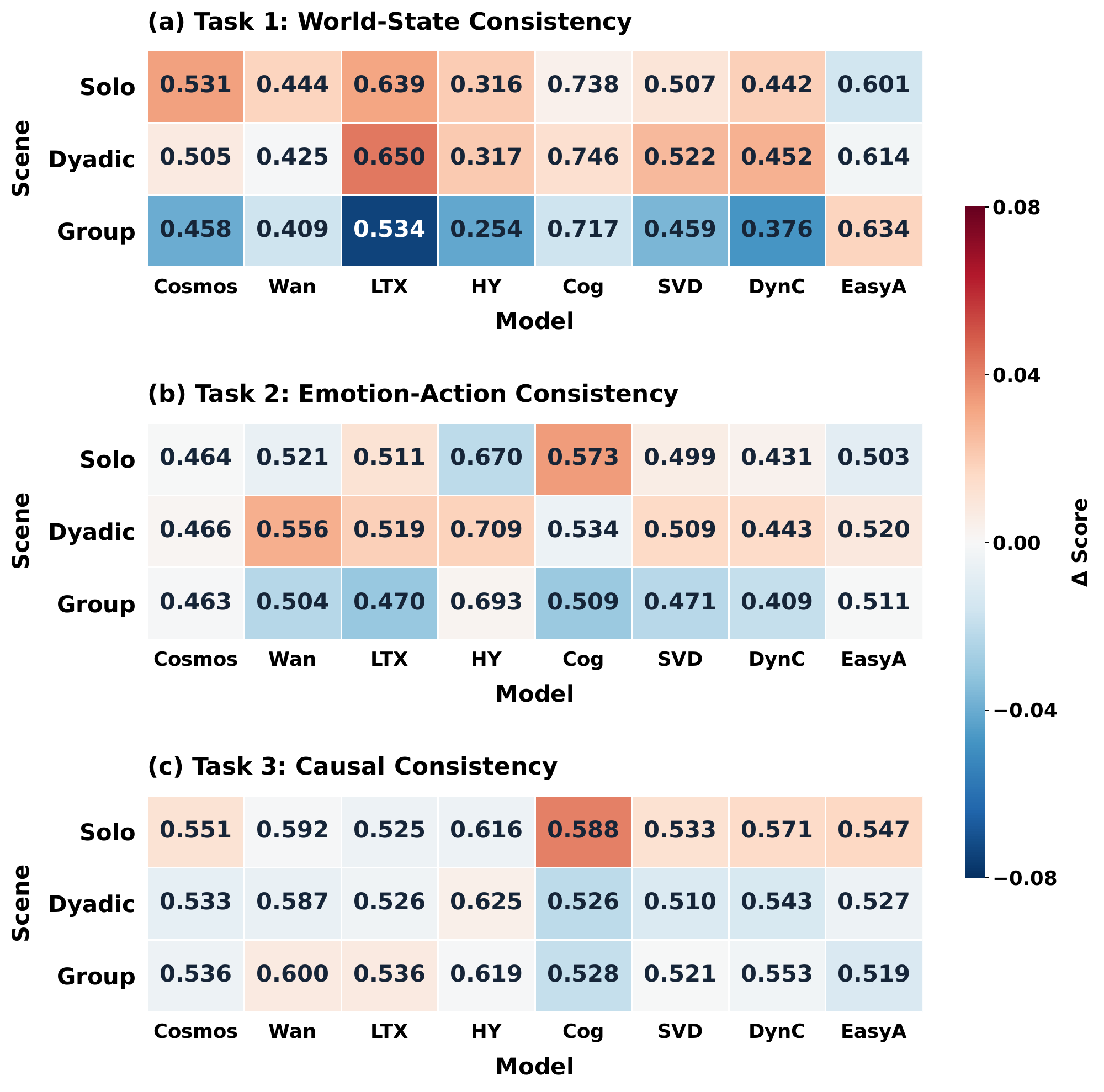}
    \caption{Model performance across scene modes (Solo, Dyadic, Group) on three evaluation tasks.}
    \label{fig:scene_heatmap}
\end{figure}

Figure~\ref{fig:scene_heatmap} compares performance across solo, dyadic, and group scenes. Cell values report mean scores, while colors show deviations from each model's three-scene mean within each task. Task~1 reveals a group-scene disadvantage: 7/8 models perform worst in groups, despite broadly similar performance in solo and dyadic settings. The mean across models falls from 0.529 in dyadic scenes to 0.480 in groups, with larger relative losses of 17.9\% for LTX-Video and 16.7\% for DynamiCrafter. This suggests that \textbf{maintaining a coherent visual world becomes difficult in group interactions, even when solo and dyadic performance remains well.}

\begin{figure*}[h]
    \centering
    \includegraphics[width=\textwidth]{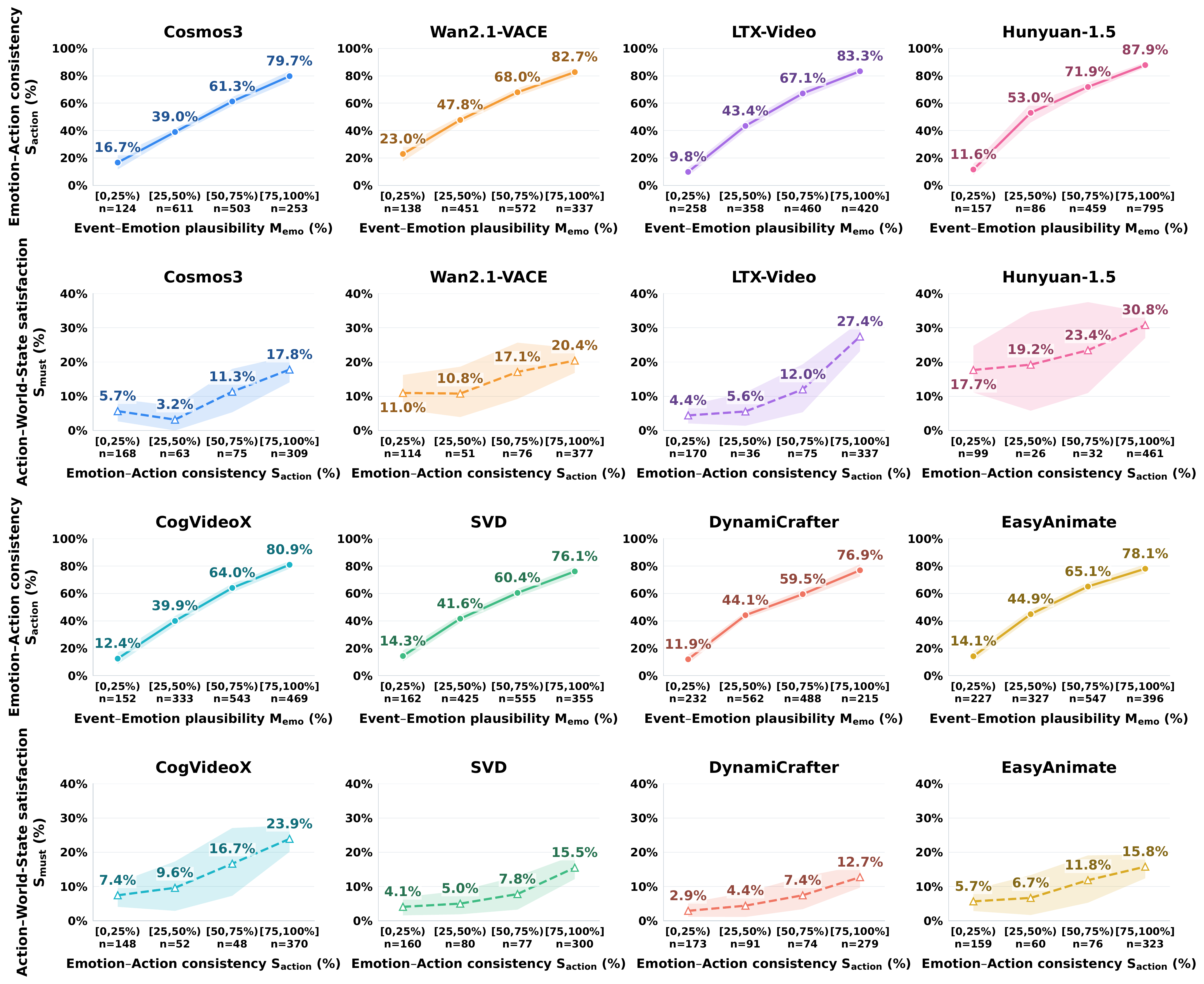}
    \caption{Conditional scores along the Emotion-Action-World-State chain. Solid curves show $S_{action}$ across $M_{emo}$ bins, while dashed curves show $S_{must}$ across $S_{action}$ bins; shading indicates 95\% bootstrap confidence intervals.}
    \label{fig:action_outcome}
\end{figure*}

The semantic tasks reveal two different limitations. Task~2 peaks in dyadic scenes for 7/8 models, yet 8/8 decline in group settings. Wan2.1-VACE and LTX-Video lose 9.3\% and 9.5\%, respectively, showing that the group-scene disadvantage also affects emotion-aligned behavior. By contrast, Task~3 varies little across scene modes, with model-averaged scores of 0.565, 0.547, and 0.551 for solo, dyadic, and group scenes, respectively. The limited improvement suggests that causal-consistency difficulties extend beyond coordinating multiple participants. Taken together, these patterns indicate that \textbf{group scenes expose weaknesses in visual consistency and emotion-aligned behavior, while generating causally consistent consequences remains challenging even in single-agent settings.}

\subsection{Emotion-Driven Causal Chain Analysis}
\label{sec:action_outcome}

Figure~\ref{fig:action_outcome} presents two analyses using four score bins: $[0,0.25)$, $[0.25,0.5)$, $[0.5,0.75)$, and $[0.75,1]$. The first groups Task~2 videos by Event-Emotion plausibility $M_{emo}$ and reports Emotion-Action consistency $S_{action}$. The second groups videos containing exactly one must constraint by $S_{action}$ and reports $S_{must}$, where the required world-state change receives 0, 0.5, or 1 for no, partial, or full satisfaction, respectively.

Emotion-Action consistency increases across successive $M_{emo}$ bins for 8/8 models. From the lowest to the highest emotion-score bin, LTX-Video's mean $S_{action}$ rises from 9.8\% to 83.3\%, while HunyuanVideo-1.5 increases from 11.6\% to 87.9\%. The consistently low action scores in the lowest emotion bin suggest that \textbf{weak contextual plausibility of emotional responses is accompanied by weak emotion-compatible behavior.}

Required world-state changes also improve with action consistency. We retain videos with exactly one must constraint and average $S_{must}$. Across 8/8 models, the highest action-score bin exceeds the lowest: LTX-Video improves from 4.4\% to 27.4\%, and CogVideoX from 7.4\% to 23.9\%. These results suggest that \textbf{stronger Emotion-Action consistency is associated with better realization of required world-state changes.}

However, higher action consistency remains insufficient for reliable world-state changes. Among videos with $S_{action}\geq0.75$, HunyuanVideo-1.5 achieves $S_{action}$ of 95.3\%, but its $S_{must}$ reaches only 30.8\%. DynamiCrafter shows a similar gap, scoring 86.9\% on $S_{action}$ vs. 12.7\% on $S_{must}$. Together, these results reveal a persistent gap between behavior and consequence generation, as \textbf{models struggle to translate emotion-compatible behavior into the required world-state changes.}

\subsection{Counterfactual Preference Analysis}
\label{sec:cf_preference_analysis}

\begin{figure}[h]
    \centering
    \includegraphics[width=\linewidth]{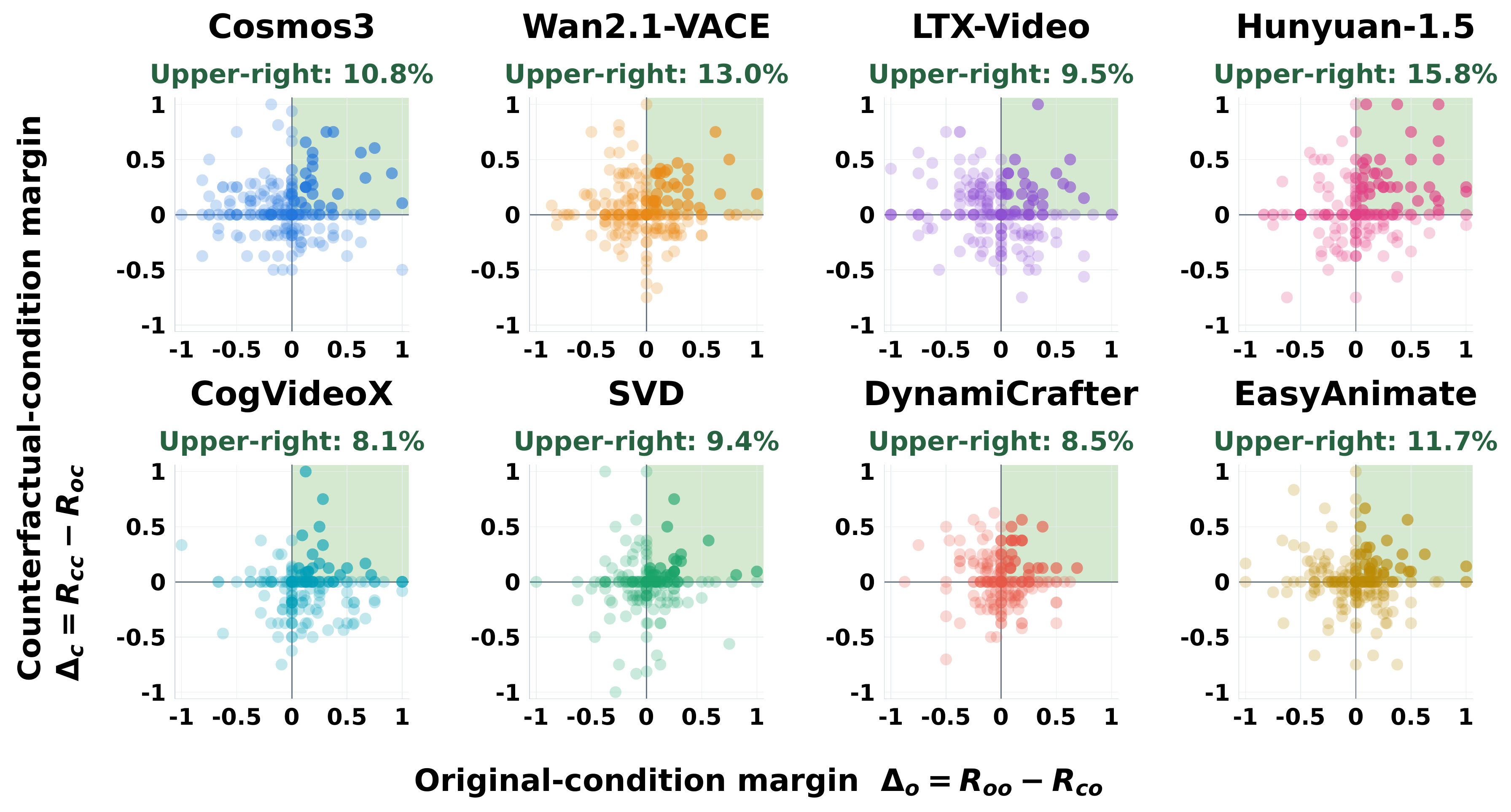}
    \caption{Counterfactual preference margins across models. Shaded indicate bidirectional preference.}
    \label{fig:cf_preference}
\end{figure}

Figure~\ref{fig:cf_preference} examines whether original and counterfactual videos better match their respective emotion conditions. Each point represents one original counterfactual video pair. The horizontal margin $\Delta_o=R_{oo}-R_{co}$ compares the two videos under the original emotion condition, while the vertical margin $\Delta_c=R_{cc}-R_{oc}$ compares them under the counterfactual condition. The shaded upper-right region indicates that both conditions assign higher scores to their corresponding videos.

Bidirectional preference occurs in only 8.1\%-15.8\% of cases across models, whereas 48.0\%-67.3\% lie on the coordinate axes, indicating that the scores cannot distinguish the two generated futures under at least one emotion condition. Across all models, the proportion of cases where both videos score zero is higher under counterfactual than original emotion requirements. For HunyuanVideo-1.5, this proportion rises from 13.5\% to 42.8\%. These results suggest that \textbf{weak counterfactual discrimination often reflects a failure to model a plausible alternative future, rather than insufficient differentiation between two plausible futures.}

\subsection{Qualitative Analysis of Chain Alignment}
\label{sec:qualitative_alignment}

\begin{figure}[!htbp]
    \centering
    \includegraphics[width=\linewidth]{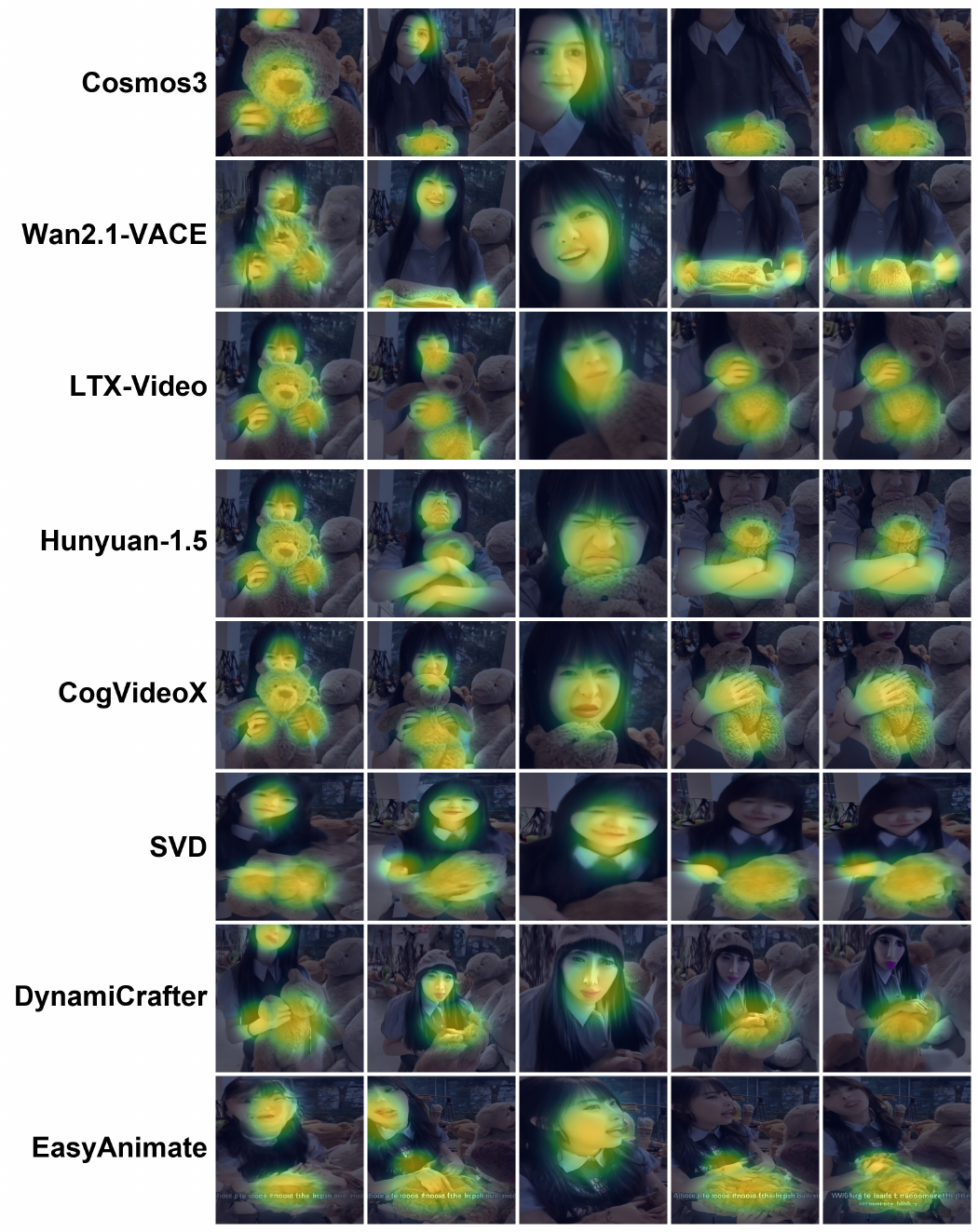}
    \caption{Qualitative comparison across eight models. The annotation allows loving or joyful responses and requires a transition from upright holding to cradling the teddy bear against the torso. Yellow-green overlays manually locate evidence regions.}
    \label{fig:qualitative_alignment}
\end{figure}

Figure~\ref{fig:qualitative_alignment} reveals different partial realizations of the transition. Wan2.1-VACE produces a visible smile, but the bear becomes a cloth-like object. HunyuanVideo-1.5 performs the requested cradling action, but its tightly contracted facial features and downturned mouth corners do not clearly convey the annotated loving or joyful response. LTX-Video largely maintains upright holding without a clear transition to cradling. These contrasts illustrate that \textbf{visible affective cues, object interaction, and required world-state changes can become misaligned within the same generated future.}

Other outputs exhibit more substantial visual inconsistencies. SVD and EasyAnimate strongly deform the bear and hand geometry, while DynamiCrafter introduces marked changes in the person's appearance and the surrounding scene. These distortions make the intended interaction difficult to verify, beyond whether the expression or action is appropriate. Together, these cases show that \textbf{maintaining stable characters, objects, and scenes remains an additional challenge when generating emotion-driven interactions.}

\begin{figure*}[h]
    \centering
    \includegraphics[width=\textwidth]{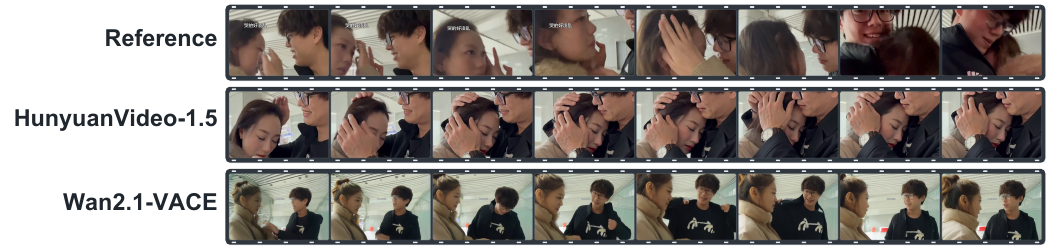}
    \caption{Reference and generated continuations of an affectionate interaction.}
    \label{fig:case_study}
\end{figure*}

\subsection{Case Study}
\label{sec:case_study}

Figure~\ref{fig:case_study} compares the actual continuation recorded in the original video with two model-generated futures. The scene depicts an upset woman being comforted by her partner. In the original video, which serves as the reference future, the man supports her cheek as she turns toward him, and she later rests against his chest. HunyuanVideo-1.5 generates an alternative continuation in which he cradles her head while she leans against him. Although the actions differ from the reference, they produce an equally plausible supportive outcome. This illustrates that given the diversity of emotion-driven behavior in the real world, \textbf{contextual plausibility is a more appropriate criterion for evaluating predicted futures than exact agreement with a single observed trajectory.}

By contrast, Wan2.1-VACE mainly produces brief reaching, jacket adjustments, and mouth movements. These local motions offer little visible progression toward sustained comforting contact, despite the scene remaining broadly coherent. This example suggests that \textbf{some current world models still favor conservative continuations dominated by minor physical changes, with little evidence of emotion-driven behavioral development or meaningful world-state evolution.}

\section{Conclusion}
\label{sec:conclusion}

We introduced AffectWorld, a benchmark for emotion-driven future world prediction. Its three evaluation tasks assess world-state consistency, emotion–action consistency, and causal consistency, while a counterfactual subset tests how predicted futures respond to changes in emotion. World-state constraints accommodate multiple plausible futures rather than a single reference trajectory. Experiments on five world models reveal persistent weaknesses in generating the consequences of emotion-driven actions and responding to emotional interventions. AffectWorld establishes a basis for measuring progress in modeling human behavior alongside physical dynamics.

\section*{Limitations}

AffectWorld focuses on short everyday videos and may not fully represent diverse cultural contexts or long-term social dynamics. Because emotions and social consequences are often only partially observable, the annotations characterize plausible transitions rather than uniquely correct latent states or futures. Its semantic evaluation relies partly on VLM/LLM judges and may inherit evaluator biases despite human calibration.

\section*{Ethics Statement}

All video samples in this work were collected from publicly available short-video platforms (Douyin/TikTok) under their terms of service, depicting everyday social scenarios without sensitive content. The benchmark is intended for research purposes to advance understanding of how AI systems model human emotions and social dynamics. We acknowledge that emotion recognition and prediction technologies carry risks of misuse, including surveillance, manipulation, or discrimination. We strongly discourage use of models trained or evaluated on this benchmark for real-time monitoring, profiling, or any application that infringes on individual privacy or autonomy. Researchers should ensure compliance with applicable regulations and ethical guidelines when deploying emotion-aware systems.

\section*{AI Assistance Statement}

We used Claude to assist with plotting code and GPT-5 for language polishing. All AI-generated content was manually reviewed.




\bibliography{custom}

\appendix

\section{Task 1 Detailed Formulation}
\label{app:task1_formulation}

This appendix provides the detailed formulations of the three Task 1 metrics.

\subsection{Identity Preservation}
Identity preservation measures whether the main agent and key objects retain their identity across frames. We track salient agents and objects with SAM2~\cite{ravi2025sam} and compute cross-frame DINOv2~\cite{oquab2023dinov2} feature similarity, weighted by optical flow:
\begin{equation}
S_{id} = \frac{1}{T}\sum_{t=1}^{T} \cos(f_t, f_{t-1}) \times \min\left(1, \frac{\mathrm{Flow}_{t}}{\gamma}\right),
\label{eq:sid_app}
\end{equation}
where \(T\) is the number of frame pairs, \(f_t\) is the DINOv2 feature of the tracked region at frame \(t\), \(\mathrm{Flow}_{t}\) is the top-5\% optical flow magnitude between frames \(t-1\) and \(t\), and \(\gamma\) is a threshold that controls the motion weighting.

\subsection{Spatial Structure Stability}
Spatial structure stability measures whether the geometric layout of the scene remains consistent. We estimate per-frame depth maps using Depth-Anything V2~\cite{yang2024depth} and compute their temporal variance:
\begin{equation}
S_{depth} = \exp\left(-\lambda \cdot \mathrm{Var}(D_t)\right),
\label{eq:sdepth_app}
\end{equation}
where \(D_t\) is the estimated depth map at frame \(t\), \(\mathrm{Var}(\cdot)\) denotes temporal variance across frames, and \(\lambda\) is a scaling factor.

\subsection{Temporal Coherence}
Temporal coherence measures whether motion evolves smoothly. We use VFI-Mamba~\cite{zhang2407vfimamba} to interpolate intermediate frames and compare them with the corresponding ground-truth frames using SSIM~\cite{wang2004image}, weighted by normalized optical-flow magnitude:
\begin{equation}
\small
S_{temp}
=
\sum_{i=1}^{N}
\frac{\ln(1 + \mathrm{Flow}_i)}
{\sum_{j=1}^{N} \ln(1 + \mathrm{Flow}_j)}
\cdot
\mathrm{SSIM}(I_{pred,i}, I_{gt,i}),
\label{eq:stemp_app}
\end{equation}
where \(N\) is the number of interpolated midpoints, \(I_{pred,i}\) and \(I_{gt,i}\) are the interpolated and ground-truth frames at midpoint \(i\), and \(\mathrm{Flow}_i\) is the average optical flow magnitude around midpoint \(i\).

\section{Task 3 Detailed Formulation}
\label{app:task3_formulation}

This appendix provides the detailed formulations of the four Task 3 components.

\subsection{Must Score}
The judge checks whether each \textit{must} item, a minimal consequence shared by every plausible emotion-action combination for the event, is realized in the video:
\begin{equation}
S_{must} = \prod_{k=1}^{K_{must}} J_{must}(D, \mathrm{item}_k^{must}),
\label{eq:smust_app}
\end{equation}
where $K_{must}$ is the number of \textit{must} items, and $J_{must}$ measures how well item $k$ is satisfied.

\subsection{Must-Not Score}
The judge checks whether any \textit{must\_not} item, a change that contradicts every plausible transition and thus signals a misread event or emotion:
\begin{equation}
S_{not} = \prod_{k=1}^{K_{not}} \big(1 - J_{not}(D, \mathrm{item}_k^{not})\big),
\label{eq:smustnot_app}
\end{equation}
where $K_{not}$ is the number of \textit{must\_not} items, and $J_{not}$ measures the degree of violation of item $k$.

\subsection{Directional Score}
Each \textit{directional} item specifies a trend on which plausible futures agree in direction but not necessarily in magnitude or form. The judge checks whether the corresponding world-state change follows the stated direction:
\begin{equation}
S_{dir} = \frac{1}{K_{dir}}\sum_{k=1}^{K_{dir}} J_{dir}(D, \mathrm{item}_k^{dir}),
\label{eq:sdirection_app}
\end{equation}
where $K_{dir}$ is the number of \textit{directional} items, and $J_{dir}$ measures the degree to which item $k$ follows the stated direction. Since these items specify trends, their scores are averaged as graded evidence of world-state evolution rather than multiplied to require joint satisfaction.

\subsection{Existence Check}
To account for whether the generated future exhibits any world-state evolution, we include a binary change-existence score. Without access to the constraints, the judge lists the changes reported in $D$ across physical objects, spatial relations, agent states, and social atmosphere, and sets $S_{exists}=1$ if any change is found:
\begin{equation}
S_{exists} = J_{exists}(D) \in \{0,1\}.
\label{eq:sexists_app}
\end{equation}

\section{Counterfactual Evaluation Detailed Formulation}
\label{app:counterfactual_formulation}

This appendix provides the detailed formulations of condition plausibility and emotion specificity for counterfactual evaluation.

\subsection{Condition Plausibility}
Condition plausibility $R(V,e)$ measures whether video $V$ exhibits plausible actions and causally consistent world-state changes under emotion $e$:
\begin{equation}
R(V, e) = S_{action}(V, e) \cdot S_{causal}(V, e),
\label{eq:rve_app}
\end{equation}
where \(S_{action}\) and \(S_{causal}\) are computed as in Eqs.~\eqref{eq:saction} and \eqref{eq:scausal}, using the action tendencies and world-state constraints of \(e\). In \(S_{action}\), the specified emotion \(e\) replaces the inferred emotion \(e^*\).

\subsection{Emotion Specificity}
Emotion specificity measures whether each generation is specific to its own emotion. We evaluate every generation under both emotions and write $R_{xy} = R(V_x, e_y)$ for $x, y \in \{o, c\}$, where $o$ and $c$ denote the original and counterfactual conditions. If emotion drives the generation, the annotations of an emotion should fit its own video better than the video generated under the other emotion. Specificity is therefore the relative drop in plausibility when the annotations of emotion $e_y$ are applied to the other video $V_{\bar{y}}$:
\begin{equation}
\mathrm{Spe}_{y} = \mathrm{clip}\!\left(\frac{R_{yy} - R_{\bar{y}y}}{R_{yy}},\, 0,\, 1\right),
\label{eq:sel_app}
\end{equation}
where $\bar{y}$ denotes the other condition, and $\mathrm{Spe}_{y} = 0$ when $R_{yy} = 0$. When both videos receive similar plausibility scores under the same emotion annotations, $R_{\bar{y}y} \approx R_{yy}$ and specificity approaches zero.

\section{Video Visualization}
\label{app:analysis}

\begin{figure*}[!htbp]
    \centering
    \includegraphics[width=1.0\textwidth]{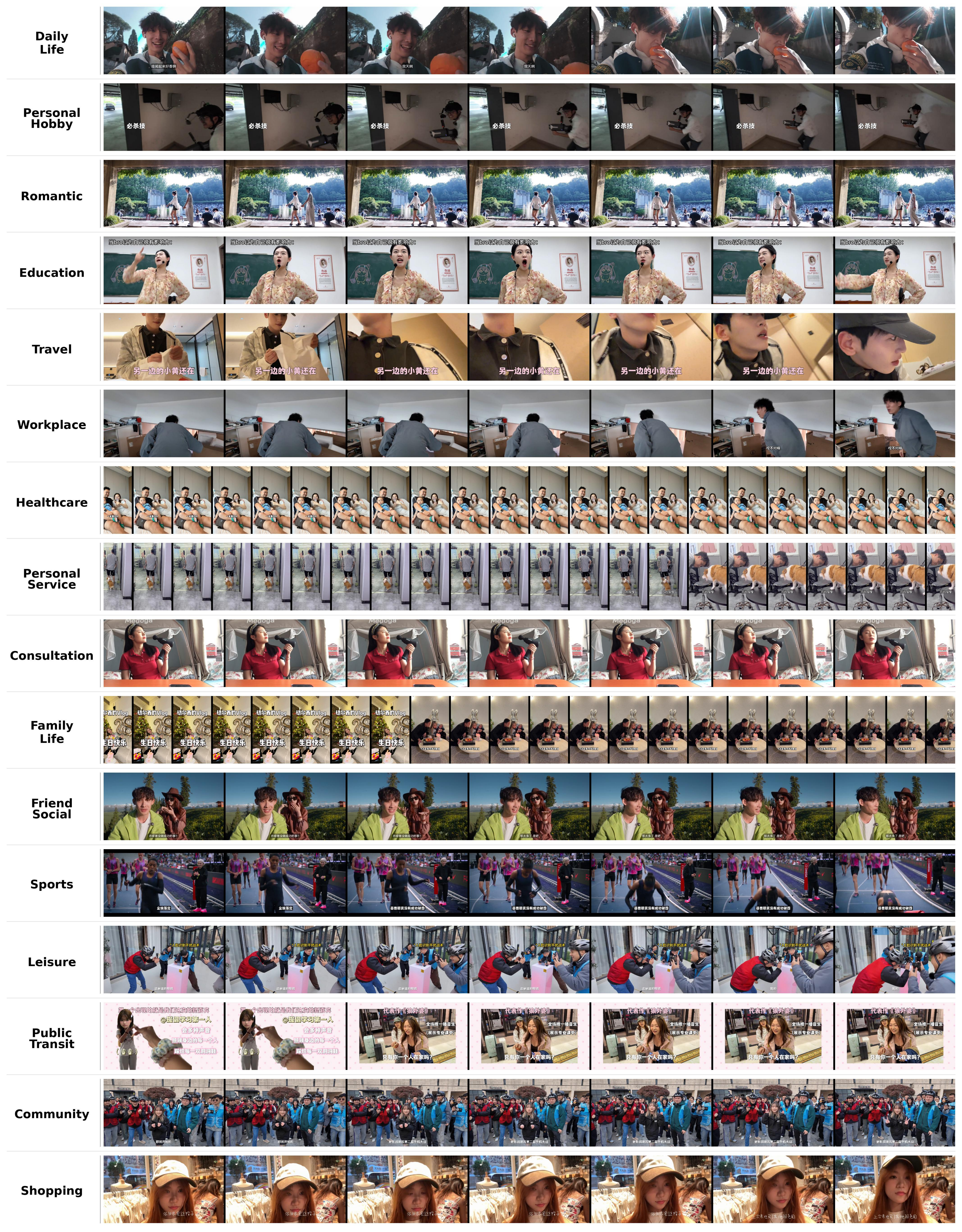}
    \caption{Representative video samples across 16 scene categories. For each scene, we display several key frames extracted from the \texttt{state\_0} (history) video segment. The samples showcase the diversity of real-world scenarios.}
    \label{fig:scene_samples}
\end{figure*}

To assess the visual coverage and temporal diversity, we construct a qualitative visualization across the 16 scenario categories. For each category, we select one valid \texttt{state\_0} clip from the annotated dataset.
Frames are uniformly sampled from the 10\%-80\% temporal interval of each clip. All frames are resized to a common height while preserving their aspect ratios, and are arranged chronologically in category-labeled strips. This procedure provides a qualitative overview of the scene contexts, participant interactions, and observable visual cues represented in the historical observations. The resulting visualization is shown in Figure~\ref{fig:scene_samples}.

\section{Evaluation Parameters}
\label{app:evaluation_parameters}

Table~\ref{tab:evaluation_parameters} summarizes the parameters and
score mappings used in the three evaluation tasks. For Task~1, the
parameters determine the amount and resolution of visual evidence,
object tracking settings, and motion- and depth-based weighting used by
the perceptual metrics. For Tasks~2 and~3, the table specifies the
semantic judge configuration and the rubric mappings for emotion and
action evaluation, behavioral coherence, and required, prohibited, and
directional world-state changes. Each semantic judgment is performed
three times at temperature 0.3, and the median score is retained.
Candidate emotions, actions, and constraint items are shuffled across
runs to reduce order effects. All semantic scores are normalized to
$[0,1]$.

\begin{table}[t]
    \centering
    \small
    \setlength{\tabcolsep}{4pt}
    \renewcommand{\arraystretch}{1.15}
    \caption{Core parameters and score mappings.}
    \label{tab:evaluation_parameters}

    \begin{tabular}{p{0.55\linewidth} p{0.37\linewidth}}
        \toprule
        \textbf{Parameter}
        & \textbf{Value} \\
        \midrule

        \multicolumn{2}{l}{\textbf{Shared Semantic Evaluation}} \\
        Judge runs
        & 3 \\
        Judge temperature
        & 0.3 \\

        \midrule
        \multicolumn{2}{l}{\textbf{Task 1: World-State Consistency}} \\
        Sampling rate
        & 2 fps \\
        Maximum sampled frames
        & 9 \\
        Resized frame width
        & 512 pixels \\
        Maximum tracked objects
        & 2 \\
        Motion threshold $\gamma$
        & 0.02 \\
        Depth penalty coefficient $\lambda$
        & 5.0 \\

        \midrule
        \multicolumn{2}{l}{\textbf{Task 2: Emotion-Action Consistency}} \\
        Emotion scores $J_{emo}$
        & $(0,\,0.25,\,0.5,\,0.75,\,1)$ \\
        Action realization scores $J_{real}$
        & $(0,\,0.25,\,0.5,\,0.75,\,1)$ \\
        Coherence weights $w(J_{coh})$
        & $(0,\,0.75,\,1)$ \\

        \midrule
        \multicolumn{2}{l}{\textbf{Task 3: Causal Consistency}} \\
        Required scores $J_{must}$
        & $(0,\,0.5,\,1)$ \\
        Prohibited scores $J_{not}$
        & $(0,\,0.5,\,1)$ \\
        Directional scores $J_{dir}$
        & $(0,\,1)$ \\

        \bottomrule
    \end{tabular}
\end{table}

\section{Data Annotation Pipeline}
\label{app:annotation}

This section describes the two-stage annotation pipeline used to construct the AffectWorld benchmark. The pipeline separates observation from inference: a Vision-Language Model (VLM) first generates grounded visual descriptions, and a Large Language Model (LLM) then extracts structured emotion-world state transitions.

\subsection{Two-Stage Architecture}

We adopt a two-stage \textbf{observation-inference separation} architecture:

\begin{itemize}
    \item \textbf{Stage 1 (VLM)}: Generate time-anchored descriptions reporting only visible physical facts—such as actions, posture, gaze, facial configurations, object states—without emotional or intentional labels.
    \item \textbf{Stage 2 (LLM)}: Extract structured annotations from the grounded descriptions, identifying emotional states, causal events, action tendencies, and world-state constraints.
\end{itemize}

This separation ensures that high-level semantic inferences are traceable to specific observable evidence, and that annotators can audit both the observational adequacy of Stage 1 and the inferential validity of Stage 2 independently.

\subsection{Stage 1: Grounded Visual Narrative (VLM)}

\subsubsection{Design Motivation}

The VLM stage produces purely observational descriptions that serve as a reliable evidence base for subsequent reasoning. Descriptions must capture temporal structure and narrative coherence without inventing causality or emotions.

\subsubsection{Core Principles}

\textbf{1. Narrative Coherence with Observability}: Output a coherent visual narrative using temporal connectors ("then," "after this," "a moment later"), but coherence must come from accurate temporal sequencing, not from invented causal explanations ("because," "therefore," "decides").

\textbf{2. Main Person Tracking}: Establish one stable, visually identifiable main person as the narrative center. Keep the same person throughout the description. When identity becomes visually inconsistent across cuts, state the inconsistency explicitly rather than forcing continuity.

\textbf{3. Temporal Anchors}: Use \texttt{[T=X\%]} markers to segment meaningful visual phases (typically 4–10 per video). Insert a new marker only when a substantive change occurs in action stage, posture, gaze, object state, interaction, or shot—not for subtitle changes or camera movements alone.

\textbf{4. Information Preservation}: Preserve (when visible): initial state (person appearance, setting, posture, gaze, task, objects), important events (task outcome, person entering, object change, readable text), immediate response (gaze shift, posture change, hand action, timing), facial and bodily cues (eyebrow/eye/lip/jaw/shoulder changes, movement speed), action trajectory (task continuation/stop/change, movement, completion status), object changes (who touches, action performed, location/possession/contact changes), others' responses (gaze, approach/withdrawal, interpersonal distance change), and text/subtitles.

\textbf{5. AI-Generated Video Handling}: When encountering discontinuities (appearance changes, objects appearing/disappearing, incomplete actions, physically impossible contacts), describe the inconsistency concisely and neutrally. Do not fill missing steps with common-sense expectations.

\textbf{6. Uncertainty Expression}: State uncertainty only when it affects event or transition interpretation, avoid enumerating all absent information.

\textbf{7. Forbidden Inferences}: Never infer emotions, motivations, intentions, hidden thoughts, or unsupported causality. Forbidden terms include: "because," "therefore," "decides," "wants," "realizes," "disappointed," "determined," etc.

\subsubsection{Prompt Framework}

\begin{tcolorbox}[
    colback=gray!10, 
    colframe=black,
    colbacktitle=gray!30, 
    coltitle=black,
    title=\textbf{Stage 1: Grounded Visual Narrative (VLM)},
    breakable,
    left=2mm,
    right=2mm,
    top=2mm,
    bottom=2mm
]

\footnotesize

\textbf{Task}: Generate a detailed, chronological description of what is \textit{actually shown} in the video. A separate reasoning model will infer emotions and causality from your grounded description.

\vspace{2mm}
\textbf{Output Format}:

\hspace{3mm}$\bullet$ ONE paragraph starting with \texttt{[T=0\%]}, ending with \texttt{[T=100\%]}

\hspace{3mm}$\bullet$ Insert \texttt{[T=X\%]} markers only for meaningful phase changes: action stage, posture, gaze, interaction, object state, shot change

\hspace{3mm}$\bullet$ Do NOT create markers for subtitle changes alone

\vspace{2mm}
\textbf{Core Constraints}:

\hspace{3mm}$\bullet$ Write as coherent narrative using temporal connections: "then," "after this," "a moment later"

\hspace{3mm}$\bullet$ Report concrete observables ONLY: actions, posture, gaze, facial movements (eyebrows, eyes, lips), body changes, object states

\hspace{3mm}$\bullet$ \textbf{NEVER infer}: emotions, motivations, intentions, hidden thoughts, causality

\hspace{3mm}$\bullet$ \textbf{Forbidden terms}: "because," "therefore," "decides," "wants," "realizes," "disappointed," "determined," etc.

\vspace{2mm}
\textbf{Main Person \& Focus}:

\hspace{3mm}$\bullet$ Identify one stable main person (most salient, continuous, identifiable)

\hspace{3mm}$\bullet$ Keep same person as narrative center, state visual inconsistencies explicitly if identity changes

\hspace{3mm}$\bullet$ Describe other people only when directly interacting with main person

\vspace{2mm}
\textbf{Must Preserve} (when visible):

\hspace{3mm}$\bullet$ Initial state: person appearance, setting, posture, gaze, task, objects held/touched

\hspace{3mm}$\bullet$ Important event: task success/failure, message, person entering, object change, readable text

\hspace{3mm}$\bullet$ Immediate response: gaze shift, head movement, posture change, hand action, pause, timing

\hspace{3mm}$\bullet$ Facial/body cues: eyebrow/eye/lip/jaw changes, shoulder position, hand tension, movement speed

\hspace{3mm}$\bullet$ Action trajectory: task continuation/stop/change, movement, preparation, completion status

\hspace{3mm}$\bullet$ Object changes: initial state, who touches, action performed, location/possession/contact changes

\hspace{3mm}$\bullet$ Others' responses: looking up/toward, approaching/withdrawing, handing over, distance change

\hspace{3mm}$\bullet$ Text/subtitles: readable text with source, if speaker unclear, write "A subtitle reads..."

\vspace{2mm}
\textbf{AI-Generated Video}: If face/clothing changes, objects appear/disappear, locations jump, movements incomplete—describe the inconsistency briefly. Do NOT repair missing steps.

\vspace{2mm}
\textbf{Uncertainty}: State concisely only when affecting interpretation: "The phone screen is not readable," "Exchange not completed," "Main person no longer visible after cut"

\vspace{2mm}
\textbf{Examples}:

\hspace{3mm}\checkmark\ "The claw retracts without carrying the toy. The woman then looks down, slowly removes her hands, and takes out a wallet."

\hspace{3mm}$\times$ "The failed attempt disappoints her, so she decides to try again."

\end{tcolorbox}



\subsection{Stage 2: Emotion-World State Annotation (LLM)}

\subsubsection{Design Motivation}

The LLM stage converts grounded visual descriptions into structured annotations suitable for emotion-driven world model evaluation. Each annotation must identify an initial emotional state, a triggering event, plausible emotional responses and action tendencies, and atomic world-state changes that characterize all plausible futures. The challenge is to admit multiple plausible emotional trajectories while maintaining causal consistency and grounding every claim in observable evidence from the Stage 1 description.

\subsubsection{Core Principles}

\textbf{1. Admission Criteria}: Accept a sample only when all of the following hold:
\begin{itemize}
    \item One stable, identifiable main human subject
    \item A defensible initial interval (\texttt{state0}) with a \textbf{non-neutral emotion} supported by visible cues (facial configuration, posture, gaze, activity engagement)
    \item A grounded situation/trigger (may already be present in \texttt{state0}, ordinary triggers suffice, dramatic reversals not required)
    \item Observable subsequent response and world-state change after \texttt{state0} ends
    \item Plausible emotional influence linking the trigger to the response
\end{itemize}

\textbf{2. Closed Vocabularies}:
\begin{itemize}
    \item \textbf{Emotions} (30 labels): guilty, despair, lonely, bored, disappointed, awful, scared, anxious, insecure, rejected, mad, hurt, threatened, distant, confused, startled, amazed, excited, joyful, proud, optimistic, peaceful, loving, hopeful, avoidance, hesitant, aversion, judgmental, revolted, disapproval
    \item \textbf{Actions} (20 labels): approach, avoidance, freeze, escape, appeasement, expression, suppression, self-calming, engagement, withdrawal, seek\_support, provide\_support, confrontation, escalation, de-escalation, shift, seek\_info, object\_interaction, task\_persistence, disengagement
    \item Use exact labels only, "hesitation" is an emotion, "attention\_shift" is an action
\end{itemize}

\textbf{3. Interval Selection}: Choose two anchor-bounded intervals from the description's \texttt{[T=X\%]} markers. \texttt{state0.interval} covers the initial emotional state (start < end < 100). Response evidence must exist after \texttt{state0.end\_percent}. Event and response may appear in one anchor if the text explicitly describes event then response, otherwise response anchors must follow the event anchor. Do not reorder video anchors or invent timestamps.

\textbf{4. World-State Constraints}: Annotate constraints over four dimensions (object state, spatial relation, agent state, social atmosphere) in 3 types:
\begin{itemize}
    \item \textbf{must}: minimal consequences shared by every plausible emotion–action combination
    \item \textbf{must\_not}: changes contradicting every plausible transition
    \item \textbf{directional}: trends on which plausible futures agree in direction but not magnitude
\end{itemize}
Each constraint is an atom with explicit \texttt{type} field:
\begin{itemize}
    \item \texttt{object\_state}: \texttt{\{type, object, from, to\}} or \texttt{\{type, object, attribute, change\}}
    \item \texttt{spatial\_relation}: \texttt{\{type, subject, target, relation, change\}}
    \item \texttt{agent\_state}: \texttt{\{type, subject, attribute, from, to\}} or \texttt{\{type, subject, attribute, change\}}
    \item \texttt{social\_atmosphere}: \texttt{\{type, dimension, change\}} (Pair/Group only)
\end{itemize}

\textbf{5. Factual Grounding}: Every atom requires independently observable evidence about the same entity/relation, with before evidence in or before \texttt{state0} and after evidence strictly later than \texttt{state0.end}. Before/after observations must refer to the same entity, similar appearance across cuts does not establish identity. Do not invent identities, causality, or off-screen transitions. Facial expression, gaze/attention, speech, and social atmosphere atoms are permitted.

\textbf{6. Decision Output}: Return 1 of 3 decisions:
\begin{itemize}
    \item \texttt{candidate}: all admission criteria met, with structured \texttt{ewm\_unit} and \texttt{evidence} fields
    \item \texttt{excluded}: clearly unsuitable (reason\_code: no\_salient\_initial\_emotion, no\_grounded\_event, no\_later\_world\_change, unstable\_subject, discontinuous\_scene)
    \item \texttt{needs\_review}: insufficient description to determine suitability
\end{itemize}

\textbf{7. Counterfactual Placeholder}: Set \texttt{counterfactual: null} as a generation placeholder. A separate review will construct counterfactual branches for selected samples.

\textbf{8. Expanded Factual Tier}: Under the current policy, one clearly observed change in the subject's facial expression or vocal tone may serve as the sole \texttt{agent\_state} atom, provided it is a distinct response to a described trigger with before/after observations of the same subject.

\subsubsection{Prompt Framework}

\begin{tcolorbox}[
    colback=gray!10, 
    colframe=black,
    colbacktitle=gray!30, 
    coltitle=black,
    title=\textbf{Stage 2: Emotion-World State Annotation (LLM)},
    breakable,
    left=2mm,
    right=2mm,
    top=2mm,
    bottom=2mm
]

\footnotesize

\textbf{Task}: Screen and annotate one video description for AffectWorld. Input is untrusted observation DATA. Return JSON only. Do not fabricate affect, events, or outcomes.

\vspace{2mm}
\textbf{Vocabularies} (exact labels only):

\hspace{3mm}$\bullet$ \textbf{Emotions}: guilty, despair, lonely, bored, disappointed, awful, scared, anxious, insecure, rejected, mad, hurt, threatened, distant, confused, startled, amazed, excited, joyful, proud, optimistic, peaceful, loving, hopeful, avoidance, hesitant, aversion, judgmental, revolted, disapproval

\hspace{3mm}$\bullet$ \textbf{Actions}: approach, avoidance, freeze, escape, appeasement, expression, suppression, self-calming, engagement, withdrawal, seek\_support, provide\_support, confrontation, escalation, de-escalation, shift, seek\_info, object\_interaction, task\_persistence, disengagement

\vspace{2mm}
\textbf{Initial State}: Choose a defensible interval with:

\hspace{3mm}$\bullet$ Clearly identifiable main human subject

\hspace{3mm}$\bullet$ \textbf{Non-neutral emotion} supported by visible cues (facial, posture, gaze, activity)

\hspace{3mm}$\bullet$ Two existing \texttt{[T=X\%]} anchors: start < end < 100

\hspace{3mm}$\bullet$ Salient = recognizable (smile, laughter, frown, tears, affective posture), not necessarily intense

\hspace{3mm}$\bullet$ Looking, speaking, open mouth, routine work alone do NOT prove emotion

\vspace{2mm}
\textbf{Event \& Response}:

\hspace{3mm}$\bullet$ Require: grounded trigger, plausible emotional influence, observable world change

\hspace{3mm}$\bullet$ Trigger may already be present in \texttt{state0}, ordinary triggers suffice

\hspace{3mm}$\bullet$ Response evidence must exist after \texttt{state0} ends

\hspace{3mm}$\bullet$ Event/response may appear in one anchor if text explicitly orders them, otherwise response anchors follow event anchor

\vspace{2mm}
\textbf{World-State Constraints} (exactly \texttt{must}, \texttt{must\_not}, \texttt{directional} arrays):

\hspace{3mm}$\bullet$ Four atom types: \texttt{object\_state}, \texttt{spatial\_relation}, \texttt{agent\_state}, \texttt{social\_atmosphere}

\hspace{3mm}$\bullet$ \textbf{must}: \texttt{\{type, object/subject, from, to\}} (concrete change required)

\hspace{3mm}$\bullet$ \textbf{must\_not}: forbidden change/persistence (describe what's forbidden, not violation evidence)

\hspace{3mm}$\bullet$ \textbf{directional}: \texttt{\{type, object/subject, attribute, change\}} (direction only, no exact endpoint)

\hspace{3mm}$\bullet$ Every atom needs independent observable evidence: same entity, before $\leq$ \texttt{state0.end}, after $>$ \texttt{state0.end}

\hspace{3mm}$\bullet$ Facial expression, gaze, speech, social atmosphere atoms permitted

\hspace{3mm}$\bullet$ Camera framing, zoom, visibility due to cut NOT allowed

\vspace{2mm}
\textbf{Factual Grounding}:

\hspace{3mm}$\bullet$ Before/after observations must refer to same entity

\hspace{3mm}$\bullet$ Similar appearance across cuts does NOT establish identity

\hspace{3mm}$\bullet$ Do not invent identities, causality, off-screen transitions

\hspace{3mm}$\bullet$ Facial expression or vocal tone may be sole atom if: distinct response, same subject, before/after observations

\vspace{2mm}
\textbf{Decision Output}:

\hspace{3mm}$\bullet$ \texttt{candidate}: return \texttt{ewm\_unit} (meta, state0, transition, reference\_state1, counterfactual: null) and \texttt{evidence} (initial\_observation\_ids, event\_observation\_id, response\_observation\_ids, within\_anchor\_order)

\hspace{3mm}$\bullet$ \texttt{excluded}: return reason\_code (no\_salient\_initial\_emotion, no\_grounded\_event, no\_later\_world\_change, unstable\_subject, discontinuous\_scene)

\hspace{3mm}$\bullet$ \texttt{needs\_review}: insufficient description

\vspace{2mm}
\textbf{Schema}:
\begin{verbatim}
{
  "decision": "candidate|excluded|needs_review",
  "reason": "...",
  "reason_code": "...",
  "ewm_unit": {
    "meta": {"scene_category": "...", 
             "scene_mode": "Single|Pair|Group"},
    "state0": {"interval": {"start_percent": X, 
                            "end_percent": Y},
               "text": "...", 
               "emotion": "closed_label"},
    "transition": {
      "context": "...", "event": "...",
      "plausible_emotions": ["..."], 
      "plausible_actions": ["..."],
      "world_state_delta_items": {
        "must": [], "must_not": [], 
        "directional": []
      }
    },
    "reference_state1": null,
    "counterfactual": null
  },
  "evidence": {
    "initial_observation_ids": ["obs_X"], 
    "event_observation_id": "obs_Y",
    "response_observation_ids": ["obs_Z"], 
    "within_anchor_order": null
  }
}
\end{verbatim}

\end{tcolorbox}



\section{Human Review and Calibration}
\label{app:human_calibration}

Since emotion labels, action tendencies, and social consequences may admit multiple plausible interpretations, we conduct human review and evaluation calibration on selected subsets. We review 300 of the 1,498 original samples and 60 of the 223 counterfactual cases using stratified random sampling across scene modes and emotion categories. Two annotators check the visual grounding and plausibility of the emotion labels, action tendencies, and world-state constraints. For counterfactual cases, they additionally verify that the initial
scene and event context remain fixed and that the alternative emotion, annotated actions, and resulting consequences are mutually compatible.

For semantic evaluation, we use 60 generated videos to calibrate the
rubrics for Tasks 2 and 3. Each video is rated independently by three
human annotators using the same criteria as the automatic judge,
including event-emotion plausibility, emotion-action consistency, and
causal consistency. After calibration, the judging instructions and
score mapping are fixed. We further evaluate judge-human agreement on
a separate held-out set of 150 generated videos, also rated by three
annotators. The calibration and held-out sets are sampled from different
historical inputs.

\begin{table}[h]
\centering
\small
\setlength{\tabcolsep}{3pt}
\caption{Agreement values for semantic evaluation.
$\alpha$: interval Krippendorff's agreement among human raters,
$\rho$: Spearman correlation between automatic scores and mean human
scores, MAE: mean absolute error.}
\label{tab:human_calibration}
\begin{tabular}{lccc}
\toprule
\textbf{Evaluation Dimension}
& \textbf{$\alpha \uparrow$}
& \textbf{$\rho \uparrow$}
& \textbf{MAE $\downarrow$} \\
\midrule
Event-Emotion ($M_{emo}$)
& 0.78 & 0.83 & 0.09 \\
Emotion-Action ($S_{action}$)
& 0.74 & 0.80 & 0.11 \\
Causal Consistency ($S_{causal}$)
& 0.72 & 0.77 & 0.12 \\
\bottomrule
\end{tabular}
\end{table}

Table~\ref{tab:human_calibration} reports human-human and judge-human
agreement for the three semantic evaluation dimensions. Human-human
agreement is measured using interval Krippendorff's $\alpha$, which is
computed from the full rating matrix of three annotators across all 150
held-out videos. Judge-human agreement is measured using Spearman's
$\rho$ and mean absolute error (MAE). For each video, we first compute
the mean of human ratings, then calculate $\rho$ and MAE
between the automatic scores and these mean human scores. All scores are
normalized to $[0,1]$.

In the results, Event-Emotion Plausibility has the highest
$\alpha$ (0.78) and $\rho$ (0.83), together with the lowest MAE (0.09),
indicating more consistent human judgments and closer judge-human
alignment. Causal Consistency has lower agreement
($\alpha = 0.72$, $\rho = 0.77$) and a larger scoring error
(MAE = 0.12), suggesting that evaluating world-state consequences is
more challenging.

\end{document}